\documentclass[10pt,twocolumn,letterpaper]{article}

\usepackage{cvpr}              

\usepackage{algorithm,algorithmic}
\usepackage{multirow}
\usepackage{makecell,enumitem}

\usepackage{amsmath}
\usepackage{amssymb}
\usepackage{mathtools}
\usepackage{amsthm}
\allowdisplaybreaks
\usepackage{subcaption}
\newtheorem{theorem}{Theorem}

\definecolor{cvprblue}{rgb}{0.21,0.49,0.74}
\usepackage[pagebackref,breaklinks,colorlinks,allcolors=cvprblue]{hyperref}

\def\paperID{10348} 
\def\confName{CVPR}
\def\confYear{2025}

\title{Unveiling Differences in Generative Models: \\ A Scalable Differential Clustering Approach}

\author{Jingwei Zhang \hspace{10pt} Mohammad Jalali \hspace{10pt} Cheuk Ting Li \hspace{10pt} Farzan Farnia\\
The Chinese University of Hong Kong\\
{\tt\small \{jwzhang22, mjalali24, farnia\}@cse.cuhk.edu.hk, ctli@ie.cuhk.edu.hk}
}
\begin{document}
\maketitle
\begin{abstract}
A fine-grained comparison of generative models requires the identification of sample types generated differently by each of the involved models. While quantitative scores have been proposed in the literature to rank different generative models, score-based evaluation and ranking do not reveal the nuanced differences between the generative models in producing different sample types. In this work, we propose solving a \emph{differential clustering} problem to detect sample types generated differently by two generative models. To solve the differential clustering problem, we develop a spectral method called \emph{Fourier-based Identification of Novel Clusters (FINC)} to identify modes produced by a generative model with a higher frequency in comparison to a reference distribution. FINC provides a scalable algorithm based on random Fourier features to estimate the eigenspace of kernel covariance matrices of two generative models and utilize the principal eigendirections to detect the sample types present more dominantly in each model. We demonstrate the application of the FINC method to large-scale computer vision datasets and generative modeling frameworks. Our numerical results suggest the scalability of the developed Fourier-based method in highlighting the sample types produced with different frequencies by generative models. The project code is available at \url{https://github.com/buyeah1109/FINC}   
\end{abstract}

\section{Introduction}
\label{sec:intro}
Deep generative models trained by variational autoencoders \cite{kingma2013auto}, generative adversarial networks \cite{goodfellow2014generative}, and denoising diffusion models \cite{ho2020denoising} have led to outstanding results on various computer vision, audio, and video datasets. To compare different generative models and rank their performance, several quantitative scores have been proposed in the computer vision community. Metrics such as Inception Score (IS) \cite{Salimans2016} and Fréchet Inception Distance (FID) \cite{heusel2017gans} have been widely used in the community for comparing the performance of modern generative model frameworks.

While the scores in the literature provide insightful evaluations of generative models, such score-based assessments do not provide a nuanced comparison between the models, which may lead to inconsistent rankings compared to human evaluations. The recent study \cite{stein2023exposing} demonstrated the imperfections of quantitative evaluation scores, showing that their rankings may not align with human-based assessments. These findings highlight the need for more detailed comparisons of generative models' produced data and the reference datasets. Specifically, a comprehensive comparison that identifies the sample types expressed differently by the models could be useful in several applications. Such a comparison can reveal suboptimally generated sample types, which could be used to further improve the performance of generative models.

To have a fine-grained comparison between two generative models, we propose solving a \emph{differential clustering} problem. This approach seeks to identify the novel sample clusters of a generative model that are generated more frequently compared to a reference distribution. By addressing the differential clustering task, we can identify the differently expressed sample types between the two models and perform an interpretable comparison revealing the types of images captured with a higher frequency by each generative model. 

To solve the differential clustering problem, one can utilize the scores proposed in the literature to quantify the uncommonness of generated data. As one such novelty score, \cite{han2023rarity} proposes the sample-based \emph{Rarity score} measuring the distance of a generated sample to the manifold of a reference distribution. In another work, \cite{jiralerspong2023feature} formulates \emph{Feature Likelihood Divergence (FLD)} as a sample-based novelty score based on the sample's likelihood according to the reference distribution. Note that the application of Rarity and FLD scores for differential clustering requires a post-clustering of the identified novel samples.
Also, \cite{zhang2024interpretable} suggests the spectral \emph{Kernel-based Entropic Novelty (KEN)} framework to detect novel sample clusters by the eigendecomposition of a kernel similarity matrix.

The discussed novelty score-based methods can potentially find the sample types differently captured by two generative models. However, the application of these methods to large-scale datasets containing many sample types, e.g. ImageNet \cite{deng2009imagenet}, will be computationally challenging. The discussed Rarity and FLD-based methods require a post-clustering of the identified novel samples which will be expensive under a relatively large number of clusters. Furthermore, the KEN method involves the eigendecomposition of a $2n\times 2n$ kernel matrix for a generated data size $n$. However, on the ImageNet dataset with hundreds of image categories, one should perform the analysis on a large sample size $n$ under which performing the eigendecomposition is computationally costly.

In this work, we develop a scalable algorithm, called \emph{Fourier-based Identification of Novel Clusters (FINC)}, to efficiently address the differential clustering problem and find novel sample types of a test generative model compared to another reference model. 
In the development of FINC, we follow the random Fourier feature framework proposed in \cite{rahimi2007random} and perform the spectral analysis for finding the novel sample types on the covariance matrix characterized by the random Fourier features.    

According to the proposed FINC method, we generate a limited number $r$ of random Fourier features to approximate the Gaussian kernel similarity score. We then run a fully stochastic algorithm to estimate the $r\times r$ covariance matrices of the test and reference generative models in the space of selected Fourier features. We prove that applying the spectral decomposition to the resulting $r\times r$  covariance matrix difference will yield the  novel clusters of the test model. As a result, the complexity of  the algorithm will primarily depend on the Fourier feature size $r$ that can be selected independently of the sample size $n$.


On the theory side, we prove an approximation guarantee that a number $ r = \mathcal{O}\bigl(\frac{\log n}{\epsilon^4}\bigr)$ of random features will be sufficient to accurately approximate the principal eigenvalues of the covariance matrices within an $\epsilon$-bounded approximation error, given $n$ samples generated by the generative models. The main challenge in analyzing the kernel method in \cite{zhang2024interpretable} is the non-Hermitian structure of their kernel matrix, which makes standard eigenvalue perturbation bounds in the literature inapplicable to analyze the perturbations to the eigenspace of the kernel matrix. To address this, we provide a novel eigenvalue perturbation analysis tailored to this non-symmetric matrix, guaranteeing that the entry-level approximations provided by random Fourier features (RFF) yield accurate proxies for the eigenspace. Our analysis ensures the sub-logarithmic growth of the required Fourier feature size $r$ in the dataset size $n$, showing the scalability of FINC for identifying differently generated modes.

We numerically evaluate the performance of our proposed FINC method on large-scale image datasets and generative models. In our numerical analysis, we attempt to identify the novel modes between state-of-the-art generative models to characterize the relative strengths of each generative modeling scheme. Also, we apply FINC to find the suboptimally captured modes of generative models with respect to their target image datasets. 
The identification of novel and more-frequently generated sample clusters could be utilized to detect generated sample types with higher memorization scores and also data types generated with higher alignment scores in text-guided generative models. 
In the following, we summarize our work's main contributions:
\begin{itemize}[leftmargin=*]
    \item Proposing a comparison approach between generative models using differential clustering,
    \item Developing the FINC method to detect differently-expressed sample types by generative models,
    \item Providing approximation guarantees on the required Fourier feature size for the FINC method,
    \item Presenting the numerical results of applying FINC to compare image-based generative models.
\end{itemize}

\section{Related Work}
\label{sec:related}

\textbf{Differential Clustering and Random Fourier Features.} Several related works \cite{chitta2012efficient, rahimi2007random, tancik2020fourier, pham2013fast} explore the application of random Fourier features \cite{rahimi2007random} to clustering and unsupervised learning tasks. Specifically, \cite{chitta2012efficient} uses random Fourier features to accelerate kernel clustering by projecting data points into a low-dimension space where the clustering is performed. \cite{tancik2020fourier} demonstrates that a  Fourier feature-based mapping enhances a neural net's capability to learn high-frequency functions in low-dimensional data domains. However, these references do not study the differential clustering task in our work. Also, \cite{he2024random} studies the RFF framework with non-symmetric kernel functions, targeting a general non-symmetric kernel that does not offer our logarithmically growing bound in Theorem~\ref{Thm: Theorem 1}. 
{Also, we note that our target differential clustering task is different from the "differential clustering" task for the single-cell RNA sequencing in the computational biology literature \cite{luecken2019current,barron2018sparse,peng2020single}. Specifically, the differential clustering proposed in \cite{barron2018sparse} aims at discovering cell types in two different biological conditions using single-cell RNA sequencing data and describing how the transcriptome profile of these cell types alters as the conditions are changing. On the other hand, our differential clustering task focuses on identifying sample types differently generated by probability models.}

\textbf{Evaluation of generative models}. The assessment of deep generative models has been the subject of a large body of related works as surveyed in \cite{Borji2022}. The proposed metrics can be categorized into the following groups: 1) distance measures between the generative model and data distribution, including FID \cite{heusel2017gans} and KID \cite{binkowski2018demystifying}, 2) scores assessing the quality and diversity of generative models including Inception score \cite{Salimans2016}, precision/recall \cite{sajjadi2018assessing,kynkaanniemi2019improved}, GAN-train/GAN-test \cite{shmelkov2018good}, density/coverage \cite{naeem2020reliable}, Vendi \cite{friedman2022vendi}, RKE \cite{jalali2023information}, FKEA-Vendi \cite{ospanov2024towards}, distributed KID-avg \cite{wang2020high}, and online multi-armed bandit evaluation schemes \cite{hu2024online,rezaei2024more,hu2024optimism}, 3) measures for the train-to-test generalization evaluation of GANs \cite{jiralerspong2023feature, alaa2022faithful, meehan2020non}. 
{Specifically, \cite{friedman2022vendi, jalali2023information, ospanov2024towards} utilize the eigenspectrum of a kernel matrix to quantify diversity. Unlike our work, these works do not aim for a comparison of generative models in generating different sample types.} As for the generalizability evaluation, the mentioned references compare a likelihood \cite{jiralerspong2023feature, alaa2022faithful} or distance \cite{meehan2020non} measure of the generated data between the training and test datasets. The distribution-based novel mode identification in our work is more general than the train-to-test generalizability, since the reference distribution in our analysis can be different from the generative model's training set. 


\textbf{Novelty assessment of generative models}. The recent works \cite{han2023rarity, jiralerspong2023feature, zhang2024interpretable} study the evaluation of novelty of samples and modes produced by generative models. \cite{han2023rarity} empirically demonstrates that rare samples are distantly located from the reference data manifold, and proposes the rarity score as the distance to the nearest neighbor for quantifying the uncommonness of a sample. \cite{jiralerspong2023feature} measures the mismatch between the likelihoods evaluated for  generated samples with respect to the training set and a reference dataset. 
We remark that \cite{han2023rarity, jiralerspong2023feature} offer \emph{sample-based} novelty scores which needs to be combined with a post-clustering step to address our target differential clustering task. However, the post-clustering step would be computationally costly under a significant number of clusters. 
Also, \cite{zhang2024interpretable} proposes a spectral method for measuring the entropy of the novel modes of a generative model with respect to a reference model, which reduces to the eigendecomposition of a kernel similarity matrix. Consequently, \cite{zhang2024interpretable}'s algorithm leads to an $\mathcal{O}(n^3)$ computational complexity given $n$ generated samples, hindering the method's application to large-scale datasets. In our work, we aim to leverage random Fourier features to reduce the spectral approach's computational complexity and to provide a solution to the differential clustering task. 

\vspace{-1mm}
\section{Preliminaries}

\subsection{Sample-based Evaluation of Generative Models}

Given a generative model $G$ generating samples from $P_G$, the goal of a quantitative evaluation is to use a group of $n$ generated data $\mathbf{x}_1,\ldots ,\mathbf{x}_n \sim P_G$ to compute a score quantifying a desired property of the generated data such as visual quality, diversity, generalizability, and novelty. Note that in such a sample-based evaluation, we do not have access to the cumulative distribution function (CDF) of $P_G$ and only observe a group of $n$ samples generated by the model. As discussed in Section \ref{sec:related}, multiple scores have been proposed in the literature to assess different properties of generative models. {We note that many novelty scores require selecting a reference dataset as a baseline to evaluate the novelty of the test distribution. Similarly, in our work, we select a test-reference distribution pair to assess how the test distribution differs from the reference distribution.}


\subsection{Kernel Function and Kernel Covariance Matrix}

Consider a kernel function $k:\mathbb{R}^d\times \mathbb{R}^d\rightarrow \mathbb{R}$ assigning the  similarity score $k(\mathbf{x} , \mathbf{x}')$ to data vectors $\mathbf{x} , \mathbf{x}'\in\mathbb{R}^d$. The kernel function is assumed to satisfy the positive-semidefinite (PSD) property requiring that for every integer $n\in\mathbb{N}$ and vectors $\mathbf{x}_1,\ldots , \mathbf{x}_n \in\mathbb{R}^d$, the $n\times n$ kernel matrix $K = \bigl[k(\mathbf{x}_i,\mathbf{x}_j)\bigr]_{n\times n}$ is a PSD matrix. Throughout this paper, we commonly use the Gaussian kernel $k_{\text{\rm Gaussian}(\sigma^2)}$ which for a bandwidth parameter $\sigma^2$ is defined as:\vspace{-2mm}
\begin{equation}\label{Eq: Gaussian kernel}
    k_{\text{\rm Gaussian}(\sigma^2)}\bigl(\mathbf{x},\mathbf{x}' \bigr)\, := \, \exp\Bigl(\frac{-\Vert \mathbf{x}- \mathbf{x}' \Vert^2_2}{2\sigma^2} \Bigr) 
\end{equation}
For every kernel function $k$, there exists a feature map $\phi:\mathbb{R}^d\rightarrow \mathbb{R}^s$ such that $k(\mathbf{x},\mathbf{x}') = \phi(\mathbf{x})^\top \phi(\mathbf{x}')$ is the inner product of $\phi$-based transformations of inputs $\mathbf{x}$ and $\mathbf{x}'$. Given the feature map $\phi$ corresponding to kernel $k$, we define the empirical kernel covariance matrix $\widehat{C}_X \in \mathbb{R}^{s\times s}$:\vspace{-2mm}
\begin{equation}\label{Eq: Kernel Covariance Matrix}
    \widehat{C}_X\, := \, \frac{1}{n}\sum_{i=1}^n \phi\bigl(\mathbf{x}_i\bigr)\phi\bigl(\mathbf{x}_i\bigr)^\top
\end{equation}
It can be seen that the kernel covariance matrix $\widehat{C}_X$ shares the same eigenvalues with the normalized kernel matrix $\frac{1}{n}K = \frac{1}{n}\bigl[ k(\mathbf{x}_i,\mathbf{x}_j)\bigr]_{n\times n}$. Assuming a normalized kernel satisfying $k(\mathbf{x},\mathbf{x}) = 1$ for every $\mathbf{x}\in\mathbb{R}^d$, a condition that holds for the Gaussian kernel in \eqref{Eq: Gaussian kernel}, the eigenvalues of $\widehat{C}_X$ will satisfy the probability axioms and result in a probability model. The resulting probability model, as theoretically shown in \cite{zhang2024interpretable}, will estimate the frequencies of the major modes of a mixture distribution. 

\subsection{Shift-Invariant Kernels and Random Fourier Features}

A kernel function $k:\mathbb{R}^d\times \mathbb{R}^d\rightarrow \mathbb{R}$ is called \emph{shift-invariant} if there exists a function $\kappa : \mathbb{R}^d\rightarrow \mathbb{R}$ such that for every $\mathbf{x} ,\mathbf{x}' \in \mathbb{R}^d$ we have
$
    k\bigl( \mathbf{x} ,\mathbf{x}' \bigr) \, =\, \kappa\bigl(\mathbf{x} -\mathbf{x}'\bigr) $. 
In other words, the output of a shift-invariant kernel $k(\mathbf{x} ,\mathbf{x}')$ only depends on the difference $\mathbf{x} -\mathbf{x}'$. For example, the Gaussian kernel in \eqref{Eq: Gaussian kernel} represents a shift-invariant kernel. Bochner's theorem \cite{bochner1949lectures} proves that a function $\kappa : \mathbb{R}^d\rightarrow \mathbb{R}$ results in a shift-invariant normalized kernel $k(\mathbf{x},\mathbf{x}')= \kappa(\mathbf{x} -\mathbf{x}')$ if and only if the Fourier transform of $\kappa$ is a probability density function (PDF) taking non-negative values and integrating to $1$. Recall that the Fourier transform of $\kappa : \mathbb{R}^d\rightarrow \mathbb{R}$, which we denote by $\widehat{\kappa}: \mathbb{R}^d\rightarrow \mathbb{R}$, is defined as
\begin{equation*}
    \widehat{\kappa}(\boldsymbol{\omega}) \, :=\, \frac{1}{(2\pi)^d}\int  \kappa(\mathbf{x}) \exp\bigl(-i \boldsymbol{\omega}^\top \mathbf{x} \bigr) \mathrm{d}\mathbf{x}
\end{equation*}
For example, in the case of Gaussian kernel \eqref{Eq: Gaussian kernel} with bandwidth parameter $\sigma^2$, the Fourier transform will be the PDF of Gaussian distribution $\mathcal{N}(\mathbf{0},\frac{1}{\sigma^2}I_d)$ with zero mean vector and isotropic covaraince matrix $\frac{1}{\sigma^2} I_d$. Based on Bochner's theorem, \cite{rahimi2007random,rahimi2008uniform,sutherland2015error} suggest using $r$ random Fourier features $\boldsymbol{\omega}_1,\ldots ,\boldsymbol{\omega}_r \sim \widehat{\kappa}$ drawn independently from PDF $\widehat{\kappa}$, and  approximating the shift-invariant kernel function
$\kappa(\mathbf{x}-\mathbf{x}')$ as follows where $\phi_{r}(\mathbf{x})=\frac{1}{\sqrt{r}}\bigl[\cos(\boldsymbol{\omega}_1^\top \mathbf{x}) ,\sin(\boldsymbol{\omega}_1^\top \mathbf{x}),\ldots ,\cos(\boldsymbol{\omega}_r^\top \mathbf{x}) ,\sin(\boldsymbol{\omega}_r^\top \mathbf{x})   \bigr]$ denotes the normalized feature map characterized by the Fourier features:
$
    k\bigl( \mathbf{x} ,\mathbf{x}' \bigr) \, \approx \, \phi_{r}(\mathbf{x})^\top \phi_{r}(\mathbf{x}')
$

\section{A Scalable Algorithm for the Identification of Novel Sample Clusters}
To provide an interpretable comparison between the test generative model $P_G$ and the reference distribution $P_{\text{\rm ref}}$, we aim to find the sample clusters generated significantly more frequently by the test model $P_G$ compared to the reference model $P_{\text{\rm ref}}$. We suppose we have access to $n$ independent samples $\mathbf{x}_1,\ldots ,\mathbf{x}_n$ from $P_G$ as well as $m$ independent samples $\mathbf{y}_1,\ldots ,\mathbf{y}_m$ from $P_{\text{\rm ref}}$.
To address the comparison task, we propose solving a \emph{differential clustering} problem to find clusters of test samples that have a significantly higher likelihood according to $P_G$ than according to $P_{\text{\rm ref}}$.

To address the differential clustering task, we follow the spectral approach proposed in \cite{zhang2024interpretable}, where for a parameter $\rho\ge 1$ we define the \emph{$\rho$-conditional kernel covariance matrix} $\Lambda_{\mathbf{X}|\rho \mathbf{Y}}$ as the difference of kernel covariance matrices $\widehat{C}_\mathbf{X},\, \widehat{C}_\mathbf{Y}$ of test and reference data, respectively, as in \eqref{Eq: Kernel Covariance Matrix}: 
\begin{equation}\label{Eq: conditional-kernel-covariance-matrix}
    \Lambda_{\mathbf{X}|\rho \mathbf{Y}} \, :=\, \widehat{C}_{\mathbf{X}} - \rho\,\widehat{C}_{\mathbf{Y}} 
\end{equation}
As proven in \cite{zhang2024interpretable}, under multi-modal distributions $P_G$ and $ P_{\text{\rm ref}}$ with well-separable modes, the eigenvectors corresponding to the positive eigenvalues of the above matrix will reveal the modes of $\mathbf{X}$ which has a frequency $\rho$-times greater than the mode's frequency in $\mathbf{Y}$. We remark that the constant $\rho$ can be interpreted as the novelty threshold, which the novelty evaluation requires the detected modes of $P_G$ to possess compared to $P_{\text{\rm ref}}$. Therefore, our goal is to find the eigenvectors corresponding to the maximum eigenvalues of the conditional covariance matrix.

The reference \cite{zhang2024interpretable} applies the kernel trick to solve the problem which leads to the eigendecomposition of an $(m+n) \times (m+n)$ kernel similarity matrix which shares the same eigenvalues with $\Lambda_{\mathbf{X}|\rho \mathbf{Y}}$. However, the resulting eigendecompositon task will be computationally challenging for significantly large $m$ and $n$ sample sizes. 
Furthermore, while the eigenvectors of \cite{zhang2024interpretable}'s kernel matrix can cluster the test and reference samples, they do not offer a clustering rule applicable to fresh samples from $P_G$ which would be desired if the learner wants to find the cluster of newly-generated samples from $P_G$. 

In this work, we propose to leverage the framework of random Fourier features (RFF) \cite{rahimi2007random} to find a scalable solution to the eigendecomposition of $\Lambda_{\mathbf{X}|\rho \mathbf{Y}}$. Unlike \cite{zhang2024interpretable}, we do not utilize the kernel trick and aim to directly apply the spectral decomposition to $\Lambda_{\mathbf{X}|\rho \mathbf{Y}}$. However, under the assumed Gaussian kernel similarity measure, the dimensions of the target $\rho$-conditional kernel covariance matrix is not finite. To resolve this challenge, we approximate the eigenspace of the original $\Lambda_{\mathbf{X}|\rho \mathbf{Y}}$ with that of the covariance matrix of the random Fourier features.  

\begin{algorithm}[t]
    \caption{Fourier-based Identification of Novel Clusters (FINC)} 
    \label{algo:KEN}
    \begin{algorithmic}[1]
    \STATE\textbf{Input:} Sample sets $\{\mathbf{x}_1,\ldots,\mathbf{x}_n\}$ and $\{\mathbf{y}_1,\ldots,\mathbf{y}_m\}$,  Gaussian kernel bandwidth $\sigma^2$, parameter $\rho$, size $r$
    \vspace{1mm}
        \STATE Draw $r$ Gaussian random vectors $\boldsymbol{\omega}_1,\ldots , \boldsymbol{\omega}_r \sim \mathcal{N}\bigl(\mathbf{0}, \frac{1}{\sigma^2}I_{d\times d}\bigr)$ \vspace{1mm}

\STATE Create the map  $\widetilde{\phi}_r(\mathbf{x}) = \frac{1}{\sqrt{r}}\bigl[\cos(\boldsymbol{\omega}_1^\top\mathbf{x}),\sin(\boldsymbol{\omega}_1^\top\mathbf{x}),\ldots ,\cos(\boldsymbol{\omega}_r^\top\mathbf{x}),\sin(\boldsymbol{\omega}_r^\top\mathbf{x})  \bigr]$
        \STATE Initialize $\widetilde{C}_X , \widetilde{C}_Y = \mathbf{0}_{r \times r} $\vspace{1mm}
        \FOR{$i\in \bigl\{1,\ldots , \max\{n,m\}\bigr\}$ }
        \STATE  Update $\widetilde{C}_X \leftarrow \widetilde{C}_X + \frac{1}{n}{\widetilde{\phi}}_r(\mathbf{x}_i){\widetilde{\phi}}_r(\mathbf{x}_i)^\top$
        \STATE  Update $\widetilde{C}_Y \leftarrow \widetilde{C}_Y + \frac{1}{m}{\widetilde{\phi}}_r(\mathbf{y}_i){\widetilde{\phi}}_r(\mathbf{y}_i)^\top$
      \ENDFOR
\STATE Compute $\widetilde{\Lambda}_{\mathbf{X}|\mathbf{Y}} = \widetilde{C}_X -\rho \widetilde{C}_Y$\vspace{1mm}
        \STATE Apply spectral eigendecomposition to obtain $\widetilde{\Lambda}_{\mathbf{X}|\mathbf{Y}} = U^\top \mathrm{diag}(\boldsymbol{\lambda})U $\vspace{1mm} 
    \STATE Find the positive eigenvalues $\lambda_1,\ldots ,\lambda_t > 0$\vspace{1mm}
    \STATE \textbf{Output:} Eigenspace $\lambda_1,\ldots,\lambda_t$, $\mathbf{u}_1,\ldots ,\mathbf{u}_t$, similarity score $s_{\mathbf{u}}(\mathbf{x})=\sum_{i=1}^r {u}_{2i-1}\cos({\boldsymbol{\omega}}_i^\top\mathbf{x})+ {u}_{2i}\sin({\boldsymbol{\omega}}_i^\top\mathbf{x})$ for every sample $\mathbf{x}$ and mode represented by $\mathbf{u}$.
    \end{algorithmic}
\end{algorithm}

Specifically, assuming a Gaussian kernel $k_{\text{\rm Gaussian}(\sigma^2)}$ with bandwidth $\sigma^2$, we sample $r$ independent vectors $\boldsymbol{\omega}_1,\ldots , \boldsymbol{\omega}_r \sim \mathcal{N}(\mathbf{0},\frac{1}{\sigma^2}I_d)$ from a zero-mean Gaussian distribution with covariance matrix $\frac{1}{\sigma^2}I_d$ where $I_d$ denotes the $d$-dimensional identity matrix. Given the randomly-sampled frequency parameters, we consider the following feature map $\widetilde{\phi}_r : \mathbb{R}^d\rightarrow\mathbb{R}^{2r},$ where $\widetilde{\phi}_r(\mathbf{x})$ is defined as
\begin{equation}
    \frac{1}{\sqrt{r}}\Bigl[ \cos\bigl(\boldsymbol{\omega}_1^\top \mathbf{x}\bigr),\sin\bigl(\boldsymbol{\omega}_1^\top \mathbf{x}\bigr),\ldots , \cos\bigl(\boldsymbol{\omega}_r^\top \mathbf{x}\bigr),\sin\bigl(\boldsymbol{\omega}_r^\top \mathbf{x}\bigr)\Bigr].
\end{equation}
Then, we compute the positive eigenvalues and corresponding eigenvectors of the following RFF-based $\rho$-conditional kernel covariance matrix ${\widetilde{\Lambda}}_{\mathbf{X}|\rho \mathbf{Y}} \in\mathbb{R}^{2r\times 2r}$:
\begin{align}\label{Eq: approx-conditional-kernel-covariance-matrix}
{\widetilde{\Lambda}}_{\mathbf{X}|\rho \mathbf{Y}} &:= \widetilde{C}_{\mathbf{X}} - \rho\widetilde{C}_{\mathbf{Y}},\\\text{\rm where }\,  \widetilde{C}_{\mathbf{X}} &:= \frac{1}{n}\sum_{i=1}^n \widetilde{\phi}_r(\mathbf{x}_i)\widetilde{\phi}_r(\mathbf{x}_i)^\top, \;\;\nonumber \\\widetilde{C}_{\mathbf{Y}} &:= \frac{1}{m}\sum_{j=1}^m \widetilde{\phi}_r(\mathbf{y}_j)\widetilde{\phi}_r(\mathbf{y}_j)^\top. \nonumber
\end{align}
Therefore, instead of the original ${{\Lambda}}_{\mathbf{X}|\rho \mathbf{Y}}$ with an infinite dimension, we only need to run the spectral eigendecomposition algorithm on the $2r\times 2r$ matrix ${\widetilde{\Lambda}}_{\mathbf{X}|\rho \mathbf{Y}}$. Furthermore, the computation of both $\widetilde{C}_{\mathbf{X}}$ and $\widetilde{C}_{\mathbf{Y}}$ can be performed using a stochastic algorithm for averaging the $2r\times 2r$ sample-based matrix $\widetilde{\phi}_r(\mathbf{x}_i)\widetilde{\phi}_r(\mathbf{x}_i)^\top$'s and $\widetilde{\phi}_r(\mathbf{y}_j)\widetilde{\phi}_r(\mathbf{y}_j)^\top$'s.

Algorithm~\ref{algo:KEN}, which we call the \emph{Fourier-based Identification of Novel Clusters (FINC)}, contains the steps explained above to reach a scalable algorithm for approximating the top more-frequently generated modes of $\mathbf{X}\sim P_G$ with respect to $\mathbf{Y}\sim P_{\text{\rm ref}}$. The following theorem proves that by choosing $r= \mathcal{O}(\frac{\log(n+m) }{\epsilon^4})$, the approximation error of the Fourier-based method is bounded by $\epsilon$.

\begin{theorem}\label{Thm: Theorem 1}
Suppose $\boldsymbol{\omega}_1,\ldots , \boldsymbol{\omega}_r\sim\widehat{\kappa}$ are drawn i.i.d. according to the Fourier transform of normalized shift-invariant kernel $\kappa$. Define $r'=\min\{2r,m+n\}$ and let ${\widetilde{\lambda}}_1 \ge \cdots \ge {\widetilde{\lambda}}_{r'}$ be the sorted top-$r'$ eigenvalues of ${\widetilde{\Lambda}}_{\mathbf{X}|\rho \mathbf{Y}}$ with corresponding eigenvectors ${\widetilde{\mathbf{v}}}_1,\ldots , {\widetilde{\mathbf{v}}}_{r'}$. Similarly, define ${{\lambda}}_1 \ge \cdots \ge {{\lambda}}_{r'}$ as the sorted top-$r'$ eigenvalues of ${{\Lambda}}_{\mathbf{X}|\rho \mathbf{Y}}$. Then, for every $\delta>0$, with probability at least $1-\delta$ we will have the following approximation guarantee with $\epsilon = \bigl(2(1+\rho)\bigr)^{3/2}\sqrt[4]{\frac{\log((m+n)/{\delta} )}{r} }$:\vspace{-3mm}
\begin{align*}
    &\sum_{i=1}^{r'} \bigl( {\widetilde{\lambda}}_{i} - {\lambda}_{i} \bigr)^2 \, \le \, \epsilon^2,\quad \\ &\max_{1\le i\le r'}\, \Bigl\Vert{\Lambda}_{\mathbf{X}|\rho \mathbf{Y}}\mathbf{v}_i - {\lambda}_{i} \mathbf{v}_i \Bigr\Vert_2 \, \le \, 2(1+\rho)\epsilon ,  
\end{align*}
where for every $i$ we define $\mathbf{v}_i = \sum_{t=1}^n \bigl(\widetilde{\mathbf{v}}^\top_i\widetilde{\phi}_r(\mathbf{x}_i)\bigr) \phi(\mathbf{x}_i) + \sum_{s=1}^m \bigl(\widetilde{\mathbf{v}}^\top_i\widetilde{\phi}_r(\mathbf{y}_s)\bigr) \phi(\mathbf{y}_s)  $ as the proxy eigenvector corresponding to the given $\widetilde{\mathbf{v}}_i$. 
\end{theorem}
Theorem~\ref{Thm: Theorem 1} proves that the eigenvalues and eigenvectors of the Fourier-based approximation $\widetilde{\Lambda}_{\mathbf{X}|\rho \mathbf{Y}}$ will be $\epsilon$-close to those of the original conditional kernel covariance matrix ${\Lambda}_{\mathbf{X}|\rho \mathbf{Y}}$, conditioned that we have $r= \mathcal{O}(\frac{\log(m+n)}{\epsilon^4})$ number of random Fourier features. Therefore, for a fixed $\epsilon$, the required number of Fourier features grows only logarithmically with the test and reference sample size $m+n$. This guarantee shows that the complexity of FINC is significantly lower than the kernel method in \cite{zhang2024interpretable} applying eigendecomposition to an $(m+n)\times (m+n)$ kernel matrix.

\section{Numerical Results}
\begin{figure*}[t]
    \centering    
    \includegraphics[width=0.9\textwidth]{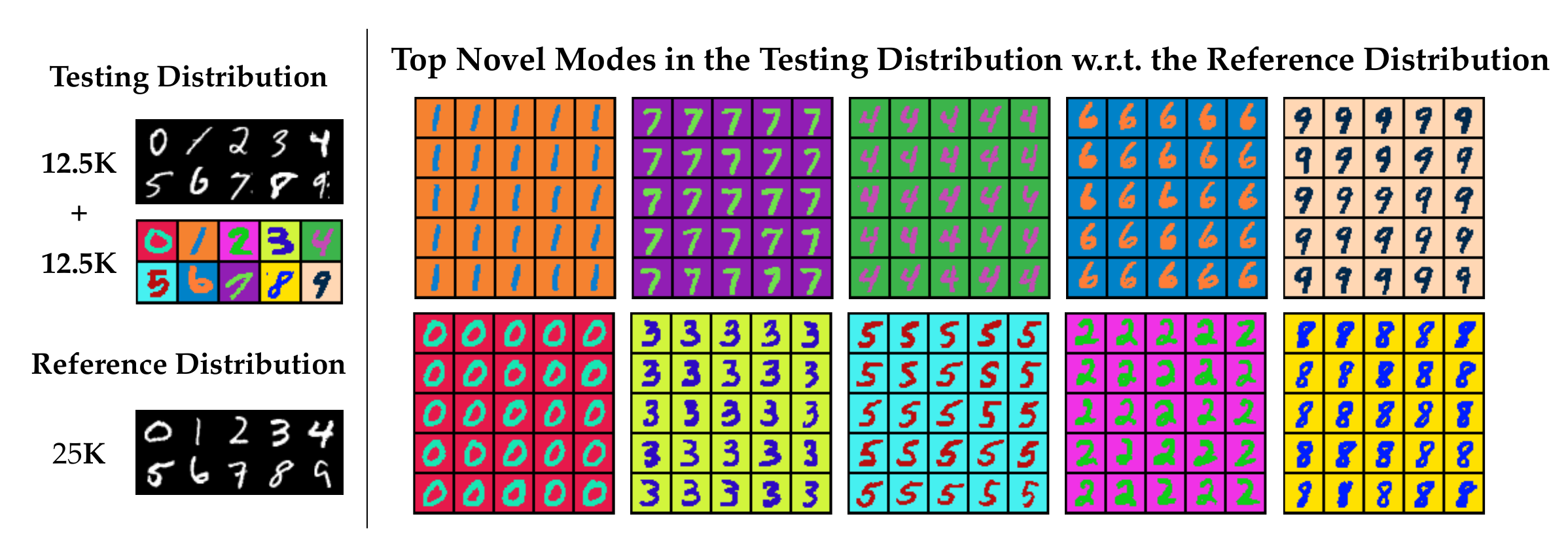}
    \vspace{-4mm}
    \caption{FINC-identified top 10 novel modes between differently-colored MNIST datasets.}\vspace{-4mm}
    \label{fig:color_mnist}  
\end{figure*}

\subsection{Experiment Setting}

\textbf{Datasets.} We performed the numerical experiments on the following standard image datasets: 1) ImageNet-1K \cite{deng2009imagenet} containing 1.4 million images with 1000 labels. The ImageNet-dogs subset used in our experiments includes 20k dog images from 120 different breeds, 2) CelebA \cite{liu2015deep} containing 200k human face images from celebrities,  3) FFHQ \cite{karras2019style} including 70k human-face images, 4) AFHQ \cite{choi2020stargan} including 15k  dogs, cats, and wildlife animal-face images. 
\\
\textbf{Feature extraction:} Following the standard in the literature of generative model evaluation, we utilized the DINOv2 model \cite{stein2023exposing} to perform feature extraction. \\ 
\textbf{Hyper-parameters selection and sample size:} We chose the kernel bandwidth $\sigma^2$ by searching for the smallest $\sigma$ value resulting in a variance bounded by $0.01$. Also, we conducted the experiments using at least $m, n = 50$k sample sizes for every generative model. For Fourier feature size $r$, we found by grid search that $r=2000$ fulfills the goal of approximation error estimate below 1\%.

\begin{figure*}
    \centering    
    \includegraphics[width=0.9\textwidth]{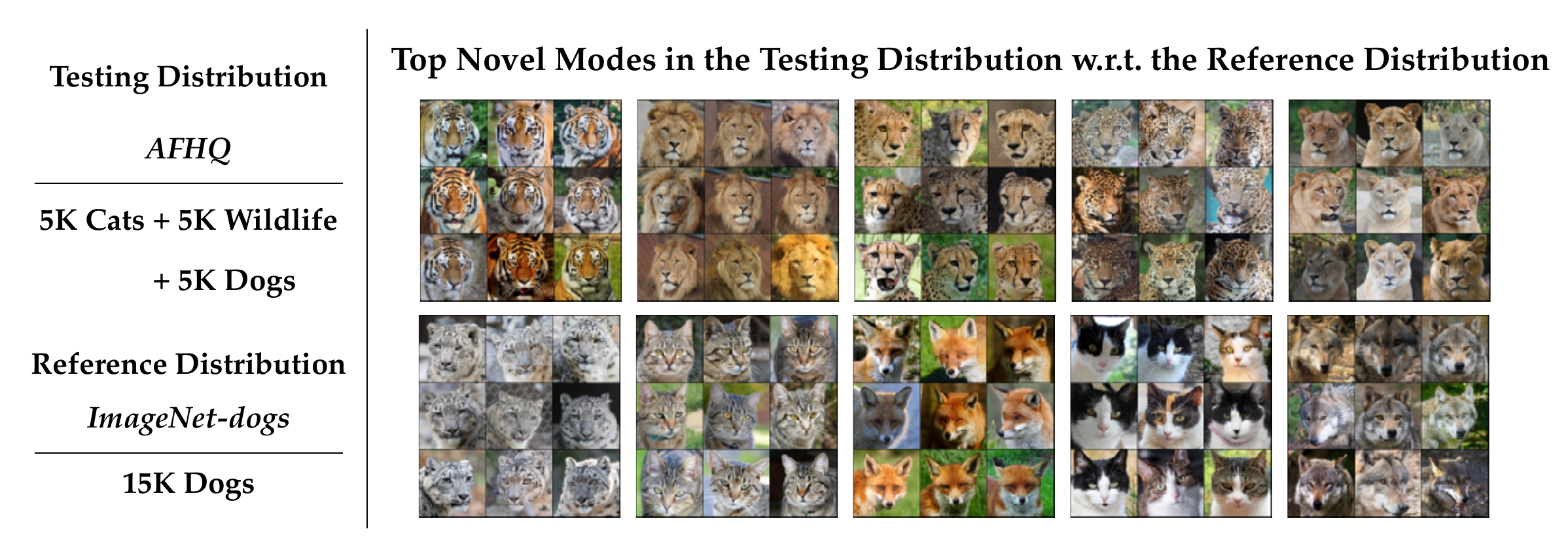}
    \vspace{-4mm}
    \caption{FINC-identified top novel modes between AFHQ \& ImageNet-dogs, DINOv2 embedding is used.}\vspace{-4mm}
    \label{fig:afhq_dogs_dinov2}  
\end{figure*}

\begin{figure*}[t]
    \centering    
    \includegraphics[width=0.85\textwidth]{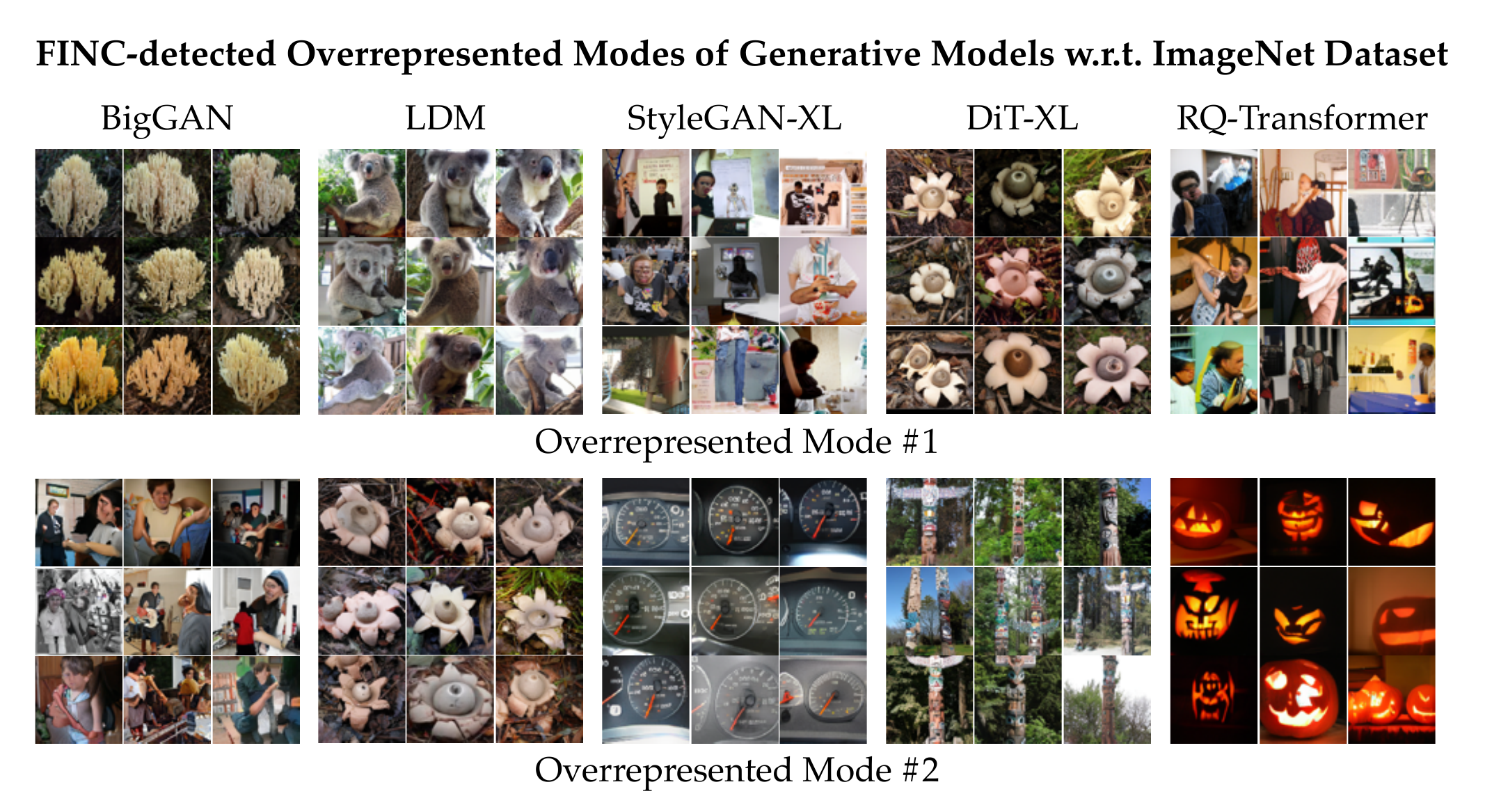}
    \vspace{-4mm}
    \caption{The FINC-identified top two overrepresented modes expressed by test generative models with a higher frequency than in the reference ImageNet dataset. DINOv2 embedding is used.}\vspace{-4mm}
    \label{fig:overrepresented_imgnet_dinov2}  
\end{figure*}

\begin{figure*}[t]
    \centering     \includegraphics[width=0.85\textwidth]{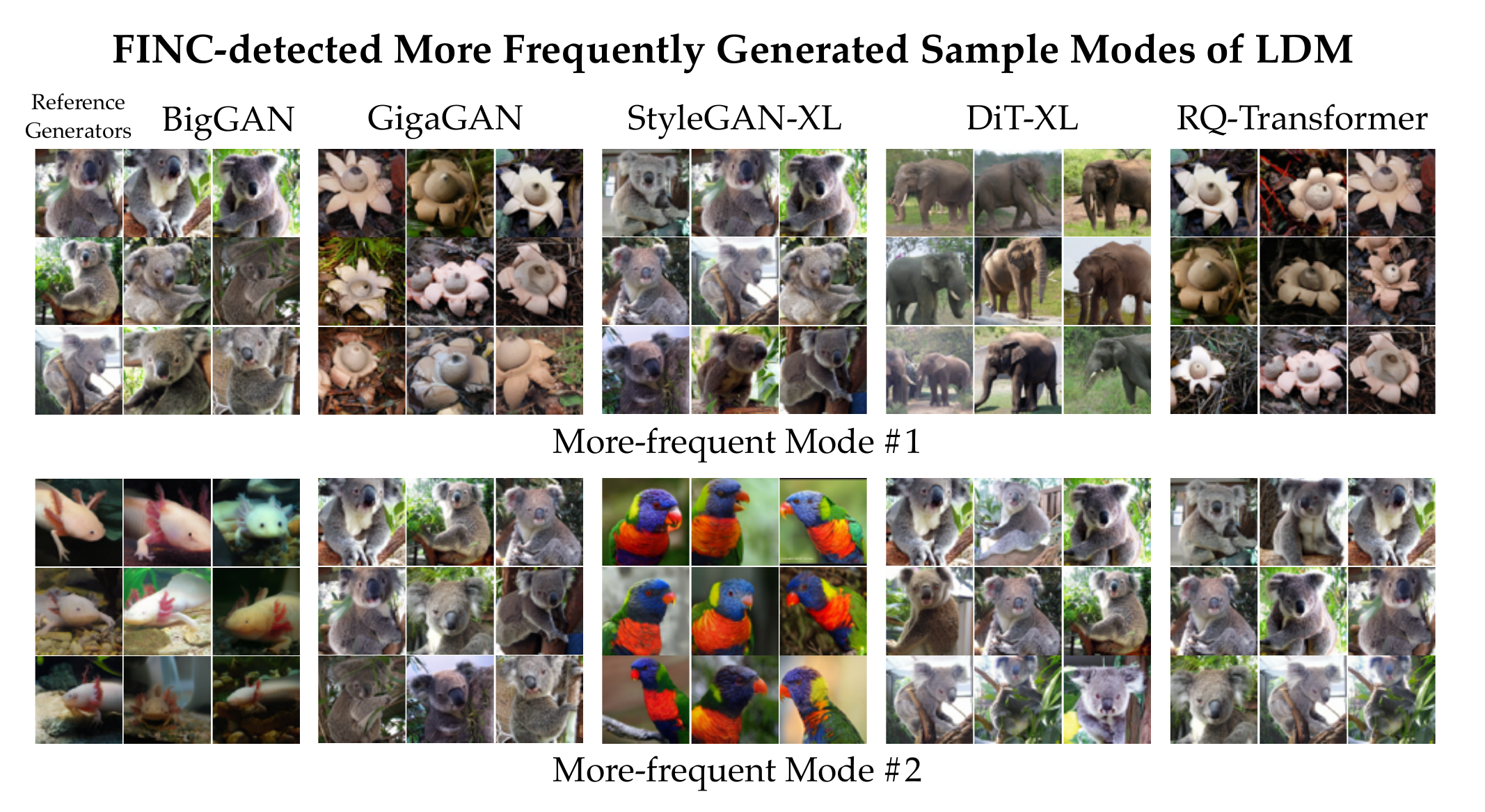}
    \vspace{-4mm}
    \caption{Representative samples from FINC-detected top-2 modes with the maximum frequency gap between LDM (higher frequency) and reference generative models. DINOv2 embedding is used.}\vspace{-4mm}
    \label{fig:higher_freq_ldm_dinov2}  
\end{figure*}

\begin{table*}[t]
\caption{Time complexity (in seconds) of solving the differential clustering problem on the ImageNet dataset using the proposed FINC method vs. the baselines. The dash - means failure due to memory overflow, and $>24h$ implies not complete in 24 hours. For baseline Rarity and FLD, post-clustering method spectral clustering is adopted. For KEN score, both eigen and Cholesky decomposition are used.} \vspace{-2mm}
\label{tab:sample_size}
\centering
\small
\begin{tabular}{clrrrrrrrr}
\toprule
                &                                         & \multicolumn{8}{c}{Sample Size}                                    \\
\cmidrule{3-10} 
\multirow{-2}{*}{Processor} & \multicolumn{1}{c}{\multirow{-2}{*}{Method}}         & 1k   & 2.5k  & 5k     & 10k     & 25k   & 50k  & 100k & 250k  \\
\midrule
                      & Rarity \cite{han2023rarity} & 1.4 & 9.1  & 37.6  & 153.5        & 1030      & 4616     & -     & -      \\
                      & FLD \cite{jiralerspong2023feature}        & 2.7 & 14.3  & 59.4  & 224.5        &  1408     & 6024     & -     & -      \\
                      &  KEN \cite{zhang2024interpretable}      & 3.5 & 26.3 & 122.8 & 1130 & 14354      & $>$24h     &  -    &  -     \\
                      &  KEN (Cholesky) \cite{zhang2024interpretable}                 & 2.6 & 16.1 & 61.7  & 266.0  & 2045      & 10934     &  -    &  -     \\
\multirow{-5}{*}{CPU} &  FINC ($r=2000$)         & 13.1 & 31.6  & 62.8  & 125.5   & 311.2 & 617.6     & 1248     & 3090      \\
\midrule

                      &  Rarity \cite{han2023rarity} & 0.1 & 4.3  & 10.7  & 27.6       & -     & -    & -    & -     \\
                      &  FLD \cite{jiralerspong2023feature}          & 0.2 & 3.7  & 10.8  & 25.1   & -     & -    & -    & -     \\
                      &  KEN \cite{zhang2024interpretable}       & 1.1 & 11.6 & 52.1  & 455.6  & -     & -    & -    & -     \\
                      &  KEN (Cholesky) \cite{zhang2024interpretable}                 & 0.1 & 0.7  & 3.2   & 16.7   & -     & -    & -    & -     \\
\multirow{-5}{*}{GPU} &  FINC ($r=2000$)          & 0.5 & 1.0  & 1.6   & 3.1    & 7.6  & 15.0 & 29.8 & 74.3 \\
\bottomrule
\end{tabular}
\vspace{-2mm}
\end{table*}

\begin{figure*}[t]
    \centering     \includegraphics[width=0.85\textwidth]{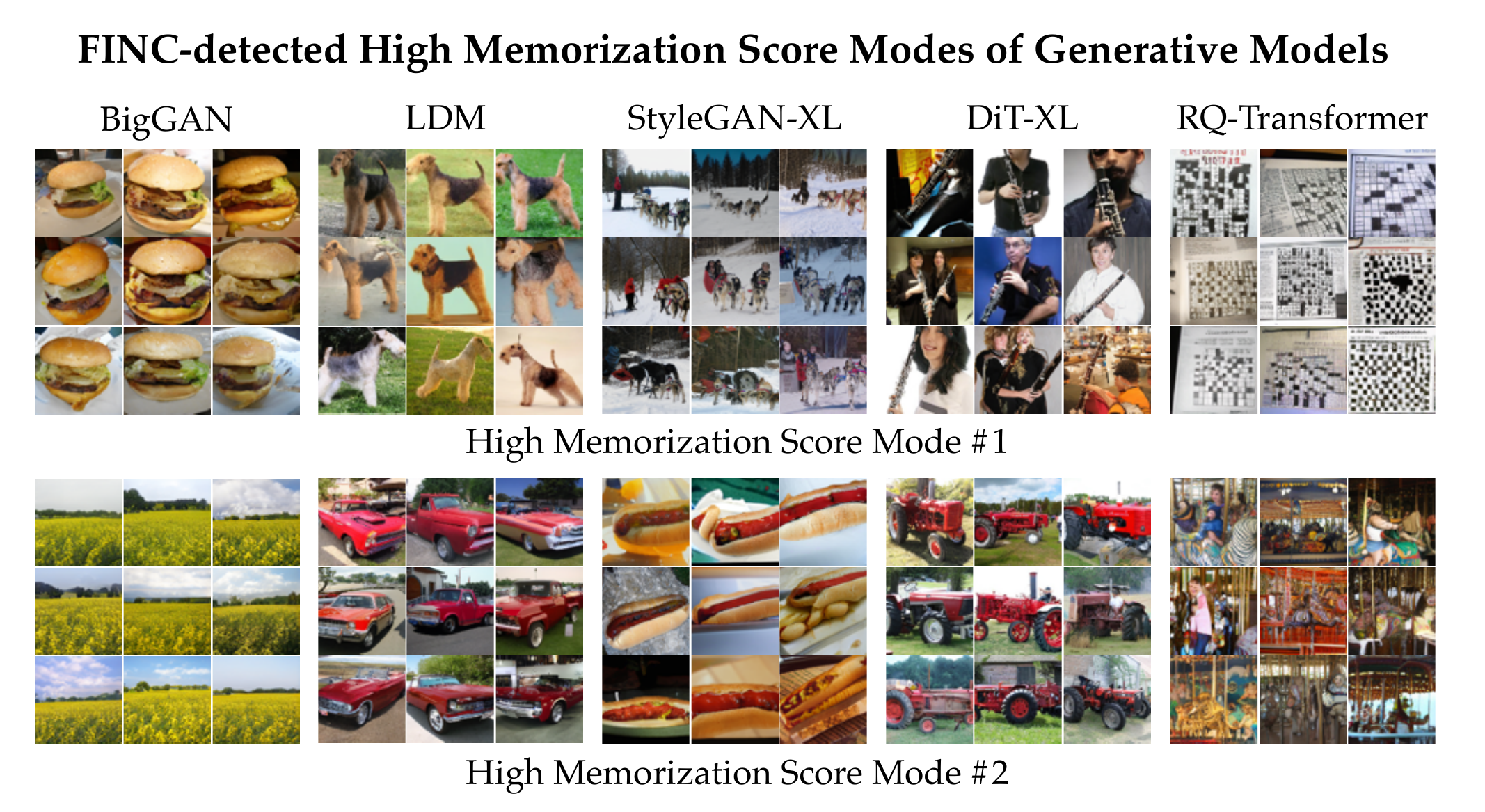}
    \vspace{-2mm}
    \caption{The FINC-identified top-2 generated sample clusters with higher FLD memorization scores w.r.t. the ImageNet training dataset. The top 9 images with the maximum alignment with the detected mode's eigenvectors are shown.}\vspace{-2mm}
    \label{fig:high_memorized_dinov2}  
\end{figure*}

\subsection{Identification of More-Frequent and Novel Modes via FINC}

\textbf{Sanity check on FINC-detected novel modes.} We empirically tested the validity of FINC's detected novel modes in controlled settings. We performed an experiment on a colored MNIST dataset. To do this, we divided the MNIST training set into two halves, one half for the reference distribution and the other for the test distribution. As a result, every grayscale \textit{sample} in the test distribution is not included in the reference dataset. Then, we colorized half of the samples in the test sample set, by assigning an exclusive color to every digit. Consequently, colored samples form the ground-truth novel sample clusters of the test distribution compared to the grayscale reference sample set. We tested whether the FINC detected top-10 novel modes match the known ground-truth novelty.

As shown in Figure~\ref{fig:color_mnist}, FINC could detect the ten types of colored digits as its top-10 novel modes. We also repeated experiments on the colorized Fashion-MNIST dataset, and observed similar results which are put in the Appendix B. We note that we used the original pixel embedding in these experiments. In addition, we investigated the image dataset pair, AFHQ and ImageNet-dogs, using the DINOv2 embedding, and as shown in Figure~\ref{fig:afhq_dogs_dinov2}, observed FINC-detected novel modes are consistent with our knowledge of the ground-truth novel modes between the datasets. 

\textbf{FINC-detected novel modes in generative models.} Next, we applied FINC to detect novel modes between various generative models trained on ImageNet-1K and FFHQ. We tested generative models following standard frameworks, including GAN-based BigGAN \cite{brock2018large}, GigaGAN \cite{kang2023gigagan}, StyleGAN-XL \cite{sauer2022stylegan}, InsGen \cite{yang2021insgen}, diffusion-based LDM \cite{rombach2021highresolution}, DiT-XL \cite{Peebles2022DiT}, and VAE-based RQ-Transformer \cite{lee2022autoregressive}, VDVAE \cite{child2020very}. For generative models trained on the ImageNet training set, we first demonstrate their overrepresented modes in Figure~\ref{fig:overrepresented_imgnet_dinov2}. Then, we illustrate novel modes between generative models. Figure~\ref{fig:higher_freq_ldm_dinov2} displays FINC-detected more frequently generated modes of LDM compared to the reference generative models with threshold $\rho=1$. We observe the higher-frequency modes of LDM is moderately consistent to its overrepresented modes (e.g. mode "koala"). Meanwhile, we can reverse the role of test and reference distribution to detect underrepresented and less frequently generated modes of generative models. Due to the page limit, we defer these results as well as results similar to Figure~\ref{fig:overrepresented_imgnet_dinov2} and \ref{fig:higher_freq_ldm_dinov2} with the FFHQ dataset, video datasets, and other generative models to the Appendix B.


\textbf{FINC mode-based similarity score.} Also, we used the eigenfunctions corresponding to the computed eigenvectors to assign a mode-based similarity score to images. As Theorem~\ref{Thm: Theorem 1} implies, the mode-scoring function for the eigenvector $\widetilde{\mathbf{v}}$ is $s_{\tilde{\mathbf{v}}}(\mathbf{x})=\sum_{i=1}^r {\tilde{v}}_{2i-1}\cos({\boldsymbol{\omega}}_i^\top\mathbf{x})+ {\tilde{v}}_{2i}\sin({\boldsymbol{\omega}}_i^\top\mathbf{x})$. This likelihood quantification can apply to both train data and fresh unseen data not contained in the sample sets for running FINC. As a result, we may retrieve the samples with largest scores assigned by top eigenvectors (ranked by associated eiganvalues) to visualize top novel modes. According to our empirical study, the FINC-based assigned scores indicate a semantically meaningful ranking of the images' likelihood of belonging to the displayed modes. In the Appendix B, we illustrate the evaluated similarity scores under two pairs of datasets, where we measured and ranked a group of unseen data according to the FINC-detected modes' scoring function.


\textbf{Scalability of FINC in the Identification of Novel Modes}. Scalability is a key property of a differential clustering algorithm in application to a data distribution with a significant number of modes, such as ImageNet. To evaluate the scalability of the proposed FINC method compared to the baselines, we recorded the time (in seconds) in Table~\ref{tab:sample_size} of running the algorithm on the ImageNet-1K dataset. In addition to the computational efficiency, we also qualitatively evaluate the novel modes detected by baselines and the FINC. For baselines, we downloaded the implementation of the Rarity score \cite{han2023rarity}, FLD score \cite{jiralerspong2023feature}, and KEN method \cite{zhang2024interpretable}, and measured their time for detecting the novel samples followed by a spectral-clustering step performed by the standard TorchCluster package.

\textbf{Quantitative: }Table~\ref{tab:sample_size} shows the time taken by FINC with $r=2000$ and the baselines on different sample sizes (we suppose $n=m$), suggesting an almost linearly growing time with sample size in the case of FINC. In contrast, the baselines' spent time grew at a superlinear rate, and for sample sizes larger than 10k and 50k all the baselines ran into a memory overflow on the GPU and CPU processors. 

\textbf{Qualitative: } We utilize the largest applicable sample size in our GPU processors for baselines. According to our empirical experiments, FINC seems to better detect novel modes compared to baselines in the ImageNet experiments. We defer the detailed results to the Appendix B. 

\subsection{Detection of High Memorization-Score Modes}

{In the literature,  memorization scores have been proposed to detect memorizing training samples by generative models. Although sample-level memorization metrics have been used to quantify and rank the similarity of every generated sample to training data, these scores do not directly reveal which sample types could be memorized more frequently by a generative model. To address this task, we apply FINC to conduct differential clustering on the groups of samples with low and high memorization FLD scores.}

{To do this, we evaluated and ranked the sample-level memorization score for all the generated samples using the FLD memorization metric with respect to the ImageNet training samples. Subsequently, we separated the two groups f samples with the top 20\% and bottom 20\% of FLD memorization scores. 
We performed differential clustering via FINC between the two groups, and used the eigenvalues/eigenvectors computed by Algorithm~\ref{algo:KEN} to quantify and rank mode-level memorization scores. Figure~\ref{fig:high_memorized_dinov2} displays the top two detected sample types that occur more frequently in the high-memorization samples of the generative models.}

\subsection{Detection of Sample Clusters with Higher and Lower CLIPScores}
The CLIPScore \cite{hessel2021clipscore} is a widely-used sample-based metric to measure the alignment of the generated image by a text-to-image model. To identify the sample types with the highest and lowest CLIPScores, we ran a differential clustering experiment between the CLIPScore-based top 20\% and the bottom 20\% of samples generated by three text-to-image models, GigaGAN, SD-XL, and Kandinsky in response to text prompts from MS-COCO dataset. We show the top-3 and bottom-3 FINC-identified modes with the maximum and minimum CLIPScores in Appendix B.

\section{Conclusion}
In this work, we developed a scalable algorithm to find the differently expressed modes of two generative models. The proposed Fourier-based approach attempts to solve a differential clustering problem to find the sample clusters generated with significantly different frequencies by the test generative model and a reference distribution. We demonstrated the application of the proposed methodology in finding the suboptimally captured modes in a generative model compared to a reference distribution. An interesting future direction to this work is to extend the comparison framework to more than two generative models, where instead of pairwise comparison, one can simultaneously compare the modes generated by multiple generative models. 


\clearpage
\clearpage
\section*{Acknowledgments}
The work of Farzan Farnia is partially supported by a grant from the Research Grants Council of the Hong Kong Special Administrative Region, China, Project 14209920, and is partially supported by CUHK Direct Research Grants with CUHK Project No. 4055164 and 4937054. 
The work of Cheuk Ting Li is partially supported by two grants from the Research Grants Council of the Hong Kong Special Administrative
Region, China [Project No.s: CUHK 24205621 (ECS), CUHK 14209823 (GRF)].  
The authors also thank the anonymous reviewers for their constructive feedback and useful suggestions. 
{
    \small
    \bibliographystyle{ieeenat_fullname}
    \bibliography{CVPR'25/main}
}

\clearpage
\onecolumn
\appendix
\section{Proofs}
\subsection{Proof of Theorem \ref{Thm: Theorem 1}}

First, note that the Gaussian probability density function (PDF) $\mathcal{N}\bigl(\mathbf{0},\sigma^{-2}I_d\bigr)$ is proportional to the Fourier transform of the Gaussian kernel in \eqref{Eq: Gaussian kernel}, because for the corresponding shift-invariant kernel producing function $\kappa_{\sigma}(\mathbf{x})=\exp\bigl(\frac{-\Vert \mathbf{x}\Vert^2_2}{2\sigma^2}\bigr)$ we have the following Fourier transform:
\begin{align*}
  \widehat{\kappa_{\sigma}}(\boldsymbol{\omega}) \, &=\, \frac{1}{(2\pi)^d}\int \kappa_{\sigma}(\mathbf{x})\exp(-i \boldsymbol{\omega}^\top \mathbf{x}) \mathrm{d}\mathbf{x}  \\
  &=\, \frac{1}{(2\pi)^d}\Bigl({2\pi}{\sigma^{2}}\Bigr)^{d/2} \exp\Bigl(-\frac{\sigma^2 \Vert \boldsymbol{\omega} \Vert^2_2}{2}\Bigr) \\
  &=\, \underbrace{\Bigl(\frac{1}{2\pi\sigma^{-2}}\Bigr)^{d/2} \exp\Bigl(-\frac{ \Vert \boldsymbol{\omega} \Vert^2_2}{2\sigma^{-2}}\Bigr)}_{\text{\rm PDF of}\;\, \mathcal{N}(\mathbf{0},\sigma^{-2}I)}.
\end{align*}
On the other hand, according to the synthesis property of Fourier transform we can write
\begin{align*}
    k_{\text{\rm Gaussian}(\sigma^2)}\bigl(\mathbf{x},\mathbf{x}' \bigr) \, &=\, \kappa_{\sigma}\bigl(\mathbf{x} - \mathbf{x}'\bigr) \\
    &\stackrel{(a)}{=} \, \int \widehat{\kappa_{\sigma}}(\boldsymbol{\omega}) \exp\bigl(i\boldsymbol{\omega}^\top (\mathbf{x} - \mathbf{x}')\bigr)\mathrm{d}\boldsymbol{\omega} \\
    &\stackrel{(b)}{=} \,  \int \widehat{\kappa_{\sigma}}(\boldsymbol{\omega}) \cos\bigl(\boldsymbol{\omega}^\top (\mathbf{x} - \mathbf{x}')\bigr)\mathrm{d}\boldsymbol{\omega} \\
    &\stackrel{(c)}{=} \, \mathbb{E}_{\boldsymbol{\omega}\sim \mathcal{N}(\mathbf{0},\sigma^{-2}I)}\Bigl[  \cos\bigl(\boldsymbol{\omega}^\top (\mathbf{x} - \mathbf{x}')\bigr)\Bigr] \\
    &= \, \mathbb{E}_{\boldsymbol{\omega}\sim \mathcal{N}(\mathbf{0},\sigma^{-2}I)}\Bigl[  \cos\bigl(\boldsymbol{\omega}^\top\mathbf{x})\cos\bigl(\boldsymbol{\omega}^\top\mathbf{x}')  + \sin\bigl(\boldsymbol{\omega}^\top\mathbf{x})\sin\bigl(\boldsymbol{\omega}^\top\mathbf{x}') \Bigr]   
\end{align*}
In the above, (a) follows from the synthesis equation of Fourier transform. (b) uses the fact that $\widehat{\kappa_\sigma}$ is an even function, leading to a zero imaginary term in the Fourier synthesis equation. (c) use the relationship between $\widehat{\kappa_\sigma}$ and the PDF of $\mathcal{N}(\mathbf{0},\sigma^{-2}I)$. Therefore, since $\bigl\vert \cos\bigl(\boldsymbol{\omega}^\top \mathbf{y}\bigr)\bigr\vert\le 1$ holds for every $\boldsymbol{\omega}$ and $\mathbf{y}$, we can apply Hoeffding's inequality to show that for independently sampled $\boldsymbol{\omega}_1 , \ldots , \boldsymbol{\omega}_r\stackrel{\mathrm{iid}}{\sim} \mathcal{N}(\mathbf{0},\sigma^{-2}I)$ the following probabilistic bound holds:
\begin{equation*}
    \mathbb{P}\biggl( \Bigl| \frac{1}{r}\sum_{i=1}^r \cos\bigl(\boldsymbol{\omega}_i^\top (\mathbf{x} - \mathbf{x}')\bigr)  - \mathbb{E}_{\boldsymbol{\omega}\sim \mathcal{N}(\mathbf{0},\sigma^{-2}I)}\Bigl[  \cos\bigl(\boldsymbol{\omega}^\top (\mathbf{x} - \mathbf{x}')\bigr)\Bigr]  \Bigr|  \ge \epsilon  \biggr) \le 2\exp\Bigl(-\frac{r\epsilon^2}{2} \Bigr)
\end{equation*}
However, note that the identity $\cos(a-b)=\cos(a)\cos(b) + \sin(a)\sin(b)$ implies that $\frac{1}{r}\sum_{i=1}^r \cos\bigl(\boldsymbol{\omega}_i^\top (\mathbf{x} - \mathbf{x}')\bigr) \, =\, \widetilde{\phi}_r(\mathbf{x})^\top \widetilde{\phi}_r(\mathbf{x}')$ and so we can rewrite the above bound as
\begin{equation*}
    \mathbb{P}\biggl( \Bigl| \widetilde{\phi}_r(\mathbf{x})^\top \widetilde{\phi}_r(\mathbf{x}')  - k_{\text{\rm Gaussian}(\sigma^2)}(\mathbf{x},\mathbf{x}')  \Bigr|  \ge \epsilon  \biggr) \le 2\exp\Bigl(-\frac{r\epsilon^2}{2} \Bigr).
\end{equation*}
Note that both $k_{\text{\rm Gaussian}(\sigma^2)}$ and $\widetilde{k}_r(\mathbf{x},\mathbf{x}')= \widetilde{\phi}_r(\mathbf{x})^\top \widetilde{\phi}_r(\mathbf{x}')$ are normalized kernels, implying that
\begin{equation*}
     \forall \mathbf{x}\in\mathbb{R}^d:\quad  \widetilde{\phi}_r(\mathbf{x})^\top \widetilde{\phi}_r(\mathbf{x})  - k_{\text{\rm Gaussian}(\sigma^2)}(\mathbf{x},\mathbf{x})  \,  = \, 0.
\end{equation*}
In addition, the application of the union bound shows that over the test sample set $\mathbf{x}_1 ,\ldots ,\mathbf{x}_n$, we can write
\begin{equation*}
    \mathbb{P}\biggl( \max_{1\le i, j \le n}  \Bigl( \widetilde{\phi}_r(\mathbf{x}_i)^\top \widetilde{\phi}_r(\mathbf{x}_j)  - k_{\text{\rm Gaussian}(\sigma^2)}(\mathbf{x}_i,\mathbf{x}_j)  \Bigr)^2  \ge \epsilon^2  \biggr) \le 2 {n\choose 2 }\exp\Bigl(-\frac{r\epsilon^2}{2} \Bigr).
\end{equation*}
Similarly, we can show the following concerning the reference sample set $\{\mathbf{y}_1,\ldots ,\mathbf{y}_m\}$ 
\begin{align*}
    &\mathbb{P}\biggl( \max_{1\le i\le n,\, 1\le j \le m}  \Bigl( \widetilde{\phi}_r(\mathbf{x}_i)^\top \widetilde{\phi}_r(\mathbf{y}_j)  - k_{\text{\rm Gaussian}(\sigma^2)}(\mathbf{x}_i,\mathbf{y}_j)  \Bigr)^2  \ge \epsilon^2  \biggr) \le 2 mn\exp\Bigl(-\frac{r\epsilon^2}{2} \Bigr), \\
    &\mathbb{P}\biggl( \max_{1\le i, j \le m}  \Bigl( \widetilde{\phi}_r(\mathbf{y}_i)^\top \widetilde{\phi}_r(\mathbf{y}_j)  - k_{\text{\rm Gaussian}(\sigma^2)}(\mathbf{y}_i,\mathbf{y}_j)  \Bigr)^2  \ge \epsilon^2  \biggr) \le 2 {m\choose 2 }\exp\Bigl(-\frac{r\epsilon^2}{2} \Bigr)
\end{align*}
Applying the union bound, we can utilize the above inequalities to show
\begin{align}\label{Eq: Prob bound for kernels}
    \mathbb{P}\biggl( 
    &\max_{1\le i, j\le n}  \Bigl( \widetilde{\phi}_r(\mathbf{x}_i)^\top \widetilde{\phi}_r(\mathbf{x}_j)  - k_{\text{\rm Gaussian}(\sigma^2)}(\mathbf{x}_i,\mathbf{x}_j)  \Bigr)^2  \ge \epsilon^2  \nonumber\\
    \text{\rm OR}\: &\max_{1\le i,j \le m}  \Bigl( \widetilde{\phi}_r(\mathbf{y}_i)^\top \widetilde{\phi}_r(\mathbf{y}_j)  - k_{\text{\rm Gaussian}(\sigma^2)}(\mathbf{y}_i,\mathbf{y}_j)  \Bigr)^2 \ge \epsilon^2 \nonumber  \\
   \text{\rm OR}\: & \max_{1\le i\le n ,\, 1\le j \le m}  \Bigl( \widetilde{\phi}_r(\mathbf{x}_i)^\top \widetilde{\phi}_r(\mathbf{y}_j)  - k_{\text{\rm Gaussian}(\sigma^2)}(\mathbf{x}_i,\mathbf{y}_j)  \Bigr)^2  \ge \epsilon^2  \biggr) \nonumber\\
   \le \; & \Bigl(2{m\choose 2 } + 2{n\choose 2 } + 2mn\Bigr)\exp\Bigl(-\frac{r\epsilon^2}{2} \Bigr) \nonumber \\
   < \; & (m+n)^2\exp\Bigl(-\frac{r\epsilon^2}{2} \Bigr)
\end{align}
On the other hand, according to Theorem 3 from \cite{zhang2024interpretable}, we know that the non-zero eigenvalues of the $\rho$-conditional covariance matrix $\Lambda_{\mathbf{X}|\rho \mathbf{Y}}:= \widehat{C}_{\mathbf{X}} - \rho \widehat{C}_{\mathbf{Y}}$ are shared with the $\rho$-conditional kernel  matrix $$K_{\mathbf{X}|\rho \mathbf{Y}}:=\begin{bmatrix} \frac{1}{n}K_{XX} & \sqrt{\frac{\rho}{mn}}K_{XY}\vspace{2mm} \\ -\sqrt{\frac{\rho}{mn}}K^\top_{XY} & \frac{-\rho}{m}K_{YY} \end{bmatrix}$$ where the kernel function is the Gaussian kernel $k_{\text{\rm Gaussian}(\sigma^2)}$. Then, defining the diagonal matrix $D=\mathrm{diag}\bigl([\underbrace{1,\ldots , 1}_{n\,\text{\rm times}} ,\underbrace{-1,\ldots , -1}_{m\,\text{\rm times}} ] \bigr) $, we observe that
\begin{equation*}
   K_{\mathbf{X}|\rho \mathbf{Y}} \,=\, D \cdot \underbrace{\begin{bmatrix} \frac{1}{n}K_{XX} & \sqrt{\frac{\rho}{mn}}K_{XY}\vspace{2mm} \\ \sqrt{\frac{\rho}{mn}}K^\top_{XY} & \frac{\rho}{m}K_{YY} \end{bmatrix}}_{K_{\mathbf{X,\rho\mathbf{Y}}}}
\end{equation*}
where we define $K_{\mathbf{X,\rho\mathbf{Y}}}$ as the joint kernel matrix that is guaranteed to be a symmetric PSD matrix. Therefore, there exists a unique symmetric and PSD matrix $\sqrt{K_{\mathbf{X,\rho\mathbf{Y}}}}$ which is the square root of $K_{\mathbf{X,\rho\mathbf{Y}}}$. Since for every matrices $A,\, B$ where $AB$ and $BA$ are well-defined, $AB$ and $BA$ share the same non-zero eigenvalues, we conclude that $K_{\mathbf{X}|\rho \mathbf{Y}} \,=\, D K_{\mathbf{X,\rho\mathbf{Y}}}$ has the same non-zero eigenvalues as $\underline{K}= \sqrt{K_{\mathbf{X,\rho\mathbf{Y}}}}\, D\, \sqrt{K_{\mathbf{X,\rho\mathbf{Y}}}}$. As a result, $\Lambda_{\mathbf{X}|\rho\mathbf{Y}}$ shares the same non-zero eigenvalues with the defined $\underline{K}$.

We repeat the above definitions (considering the Gaussian kernel $k_{\text{\rm Gaussian}(\sigma^2)}$) for the proxy kernel function $\widetilde{k}_r(\mathbf{x},\mathbf{x}')= \widetilde{\phi}_r(\mathbf{x})^\top \widetilde{\phi}_r(\mathbf{x}')$ to obtain proxy kernel matrices $\widetilde{K}_{\mathbf{X}|\rho \mathbf{Y}}$, $\widetilde{K}_{\mathbf{X},\rho \mathbf{Y}}$, $\underline{\widetilde{K}}= \sqrt{\widetilde{K}_{\mathbf{X,\rho\mathbf{Y}}}}\, D\, \sqrt{\widetilde{K}_{\mathbf{X,\rho\mathbf{Y}}}}$. We note that Equation \eqref{Eq: Prob bound for kernels} implies that
\begin{align}
    &\mathbb{P}\Bigl( \bigl\Vert \widetilde{K}_{\mathbf{X},\rho \mathbf{Y}} - K_{\mathbf{X},\rho \mathbf{Y}}\bigr\Vert^2_F \, \ge\, n^2\frac{\epsilon^2}{n^2} + 2nm\frac{\rho\epsilon^2}{nm} + m^2\frac{\rho^2\epsilon^2}{m^2}  \Bigr) \nonumber\\
    \, \le \; &(m+n)^2\exp\Bigl(-\frac{r\epsilon^2}{2}\Bigr). \nonumber\\
    \Longrightarrow\quad&\mathbb{P}\Bigl( \bigl\Vert \widetilde{K}_{\mathbf{X},\rho \mathbf{Y}} - K_{\mathbf{X},\rho \mathbf{Y}}\bigr\Vert^2_F \, \ge\, \epsilon^2(1+\rho)^2  \Bigr) \, \le \, (m+n)^2\exp\Bigl(-\frac{r\epsilon^2}{2}\Bigr). \nonumber \\
    \Longrightarrow\quad &\mathbb{P}\Bigl( \bigl\Vert \widetilde{K}_{\mathbf{X},\rho \mathbf{Y}} - K_{\mathbf{X},\rho \mathbf{Y}}\bigr\Vert_F \, \ge\, \epsilon  \Bigr) \, \le \, (m+n)^2\exp\Bigl(-\frac{r\epsilon^2}{2(1+\rho)^2}\Bigr).
\end{align}
The above inequality will imply the following where $\Vert \cdot \Vert_2$ denotes the spectral norm
\begin{align}
\mathbb{P}\Bigl( \bigl\Vert \widetilde{K}_{\mathbf{X},\rho \mathbf{Y}} - K_{\mathbf{X},\rho \mathbf{Y}}\bigr\Vert_2 \, \ge\, \epsilon  \Bigr) \, &\le \, \mathbb{P}\Bigl( \bigl\Vert \widetilde{K}_{\mathbf{X},\rho \mathbf{Y}} - K_{\mathbf{X},\rho \mathbf{Y}}\bigr\Vert_F \, \ge\, \epsilon  \Bigr) \nonumber \\
&\le \, (m+n)^2\exp\Bigl(-\frac{r\epsilon^2}{2(1+\rho)^2}\Bigr).
\end{align}
Since for every PSD matrices $A,B \succeq \mathbf{0}$, $\bigl\Vert\sqrt{A} -\sqrt{B} \bigr\Vert_2 \le \sqrt{\Vert A - B\Vert_2}$ holds \cite{phillips1987uniform}, the above inequality also proves that
\begin{align}
\mathbb{P}\Bigl( \bigl\Vert \sqrt{\widetilde{K}_{\mathbf{X},\rho \mathbf{Y}}} - \sqrt{K_{\mathbf{X},\rho \mathbf{Y}}}\bigr\Vert_2 \, \ge\, \epsilon  \Bigr) \, & \le \,\mathbb{P}\Bigl( \sqrt{\bigl\Vert \widetilde{K}_{\mathbf{X},\rho \mathbf{Y}} - K_{\mathbf{X},\rho \mathbf{Y}}\bigr\Vert_2} \, \ge\, \epsilon  \Bigr) \, \\
&\le \, (m+n)^2\exp\Bigl(-\frac{r\epsilon^4}{2(1+\rho)^2}\Bigr).
\end{align}
Next, we note the following bound on the Frobenius norm of the difference between $\underline{K}$ and $\underline{\widetilde{K}}$:
\begin{align*}
    \Bigl\Vert \underline{\widetilde{K}} - \underline{K}\Bigr\Vert_F \: &= \; \Bigl\Vert \sqrt{\widetilde{K}_{\mathbf{X,\rho\mathbf{Y}}}}\, D\, \sqrt{\widetilde{K}_{\mathbf{X,\rho\mathbf{Y}}}} - \sqrt{{K}_{\mathbf{X,\rho\mathbf{Y}}}}\, D\, \sqrt{{K}_{\mathbf{X,\rho\mathbf{Y}}}}\Bigr\Vert_F \\
    &\stackrel{(d)}{\le} \; \Bigl\Vert \sqrt{\widetilde{K}_{\mathbf{X,\rho\mathbf{Y}}}}\, D\, \sqrt{\widetilde{K}_{\mathbf{X,\rho\mathbf{Y}}}} - \sqrt{\widetilde{K}_{\mathbf{X,\rho\mathbf{Y}}}}\, D\, \sqrt{{K}_{\mathbf{X,\rho\mathbf{Y}}}}\Bigr\Vert_F  \\
    &\quad +\Bigl\Vert \sqrt{\widetilde{K}_{\mathbf{X,\rho\mathbf{Y}}}}\, D\, \sqrt{{K}_{\mathbf{X,\rho\mathbf{Y}}}} - \sqrt{{K}_{\mathbf{X,\rho\mathbf{Y}}}}\, D\, \sqrt{{K}_{\mathbf{X,\rho\mathbf{Y}}}}\Bigr\Vert_F \\
    &\stackrel{(e)}{\le} \Bigl\Vert \sqrt{\widetilde{K}_{\mathbf{X,\rho\mathbf{Y}}}}\, D\Bigr\Vert_F\,\Bigl\Vert \sqrt{\widetilde{K}_{\mathbf{X,\rho\mathbf{Y}}}} - \sqrt{{K}_{\mathbf{X,\rho\mathbf{Y}}}}\Bigr\Vert_2 \\
    &\quad +\Bigl\Vert \sqrt{\widetilde{K}_{\mathbf{X,\rho\mathbf{Y}}}}-\sqrt{{K}_{\mathbf{X,\rho\mathbf{Y}}}}\textbf{}\Bigr\Vert_2\, \Bigl\Vert D\sqrt{{K}_{\mathbf{X,\rho\mathbf{Y}}}}\Bigr\Vert_F \\
    &\stackrel{(f)}{=} \mathrm{Tr}\Bigl(\widetilde{K}_{\mathbf{X,\rho\mathbf{Y}}}\Bigr)\,\Bigl\Vert \sqrt{\widetilde{K}_{\mathbf{X,\rho\mathbf{Y}}}} - \sqrt{{K}_{\mathbf{X,\rho\mathbf{Y}}}}\Bigr\Vert_2 \\
    &\quad +\Bigl\Vert \sqrt{\widetilde{K}_{\mathbf{X,\rho\mathbf{Y}}}}-\sqrt{{K}_{\mathbf{X,\rho\mathbf{Y}}}}\textbf{}\Bigr\Vert_2\, \mathrm{Tr}\Bigl({K}_{\mathbf{X,\rho\mathbf{Y}}}\Bigr) \\
   &\stackrel{(g)}{=}\, 2(1+\rho)\Bigl\Vert \sqrt{\widetilde{K}_{\mathbf{X,\rho\mathbf{Y}}}} - \sqrt{{K}_{\mathbf{X,\rho\mathbf{Y}}}}\Bigr\Vert_2.  
\end{align*}
Here, (d) follows from the triangle inequality for the Frobenius norm. (e) follows from the inequality $\Vert AB\Vert_F \le \min\bigl\{\Vert A\Vert_F \Vert B\Vert_2 , \Vert A\Vert_2 \Vert B\Vert_F\bigr\}$ for every square matrices $A,B\in\mathbb{R}^{d\times d}$. (f) holds because $\Vert D\sqrt{K}\Vert_F = \mathrm{Tr}(\sqrt{K}D^2\sqrt{K})=\mathrm{Tr}(K)$, and (g) comes from the fact that $\mathrm{Tr}(K_{\mathbf{X},\rho\mathbf{Y}})=1+\rho$.
Combining the above results shows the following probabilistic bound: 
\begin{align}
\mathbb{P}\Bigl( \bigl\Vert \underline{\widetilde{K}} - \underline{K}\bigr\Vert_F \, \ge\, \epsilon  \Bigr)\, &\le \, \mathbb{P}\Bigl( \bigl\Vert \sqrt{\widetilde{K}_{\mathbf{X},\rho \mathbf{Y}}} - \sqrt{K_{\mathbf{X},\rho \mathbf{Y}}}\bigr\Vert_2 \, \ge\, \frac{\epsilon}{2(1+\rho)}  \Bigr) \nonumber \\
\, &\le \, (m+n)^2\exp\Bigl(-\frac{r\epsilon^4}{32(1+\rho)^6}\Bigr).
\end{align}
According to the eigenvalue-perturbation bound in \cite{hoffman2003variation}, the following holds
\begin{equation*}
    \sqrt{\sum_{i=1}^{r'}(\widetilde{\lambda}_i - \lambda_i)^2} \,\le\, \Vert \underline{\widetilde{K}} - \underline{K}\Vert_F 
\end{equation*}
which proves that
\begin{align}
\mathbb{P}\Bigl( \sqrt{\sum_{i=1}^{r'}(\widetilde{\lambda}_i - \lambda_i)^2} \, \ge\, \epsilon  \Bigr) \, \le \, (m+n)^2\exp\Bigl(-\frac{r\epsilon^4}{4(1+\rho)^3}\Bigr)
\end{align}
Defining $\delta= (m+n)^2\exp\Bigl(-\frac{r\epsilon^4}{32(1+\rho)^6}\Bigr)$ implying that $\epsilon = \sqrt{8(1+\rho)^3}\sqrt[4]
{ \frac{\log\bigl((m+n)/\delta\bigr)}{r}}$ will then result in the following which completes the proof of the first part of the theorem: 
\begin{align}
\mathbb{P}\Bigl( \sqrt{\sum_{i=1}^{r'}(\widetilde{\lambda}_i - \lambda_i)^2} \, \ge\, \epsilon  \Bigr) \, \le \, \delta.
\end{align}
Regarding the theorem's approximation guarantee for the eigenvectors, we note that for each eigenvectors $\widetilde{\mathbf{v}}_i$ of 
the proxy conditional kernel covariance matrix $\widetilde{K}_{\mathbf{X}|\rho \mathbf{Y}}$, we can write the following considering the conditional kernel covariance matrix ${K}_{\mathbf{X}|\rho \mathbf{Y}}$ and the corresponding eigenvalue $\lambda_i$:
\begin{align*}
\Bigl\Vert {K}_{\mathbf{X}|\rho\mathbf{Y}} \widetilde{\mathbf{v}}_i - {\lambda}_i\widetilde{\mathbf{v}}_i\Bigr\Vert_2 \, &\le \, \Bigl\Vert {K}_{\mathbf{X}|\rho\mathbf{Y}} \widetilde{\mathbf{v}}_i - \tilde{\lambda}_i\widetilde{\mathbf{v}}_i\Bigr\Vert_2 + \Bigl\Vert \tilde{\lambda}_i\widetilde{\mathbf{v}}_i - {\lambda}_i\widetilde{\mathbf{v}}_i \Bigr\Vert_2 \\
    \, &= \, \Bigl\Vert {K}_{\mathbf{X}|\rho\mathbf{Y}} \widetilde{\mathbf{v}}_i - \widetilde{K}_{\mathbf{X}|\rho \mathbf{Y}}\widetilde{\mathbf{v}}_i\Bigr\Vert_2 + \bigl\vert \tilde{\lambda}_i - {\lambda}_i \bigr\vert \\
    &\le \, \Bigl\Vert {K}_{\mathbf{X}|\rho\mathbf{Y}} - \widetilde{K}_{\mathbf{X}|\rho \mathbf{Y}}\Bigr\Vert_2\Bigl\Vert\widetilde{\mathbf{v}}_i\Bigr\Vert_2 + \bigl\vert \tilde{\lambda}_i - {\lambda}_i \bigr\vert
    \\
    &= \, \Bigl\Vert {K}_{\mathbf{X}|\rho\mathbf{Y}} - \widetilde{K}_{\mathbf{X}|\rho \mathbf{Y}}\Bigr\Vert_2 + \bigl\vert \tilde{\lambda}_i - {\lambda}_i \bigr\vert \\
        &\le \, \Bigl\Vert {K}_{\mathbf{X}|\rho\mathbf{Y}} - \widetilde{K}_{\mathbf{X}|\rho \mathbf{Y}}\Bigr\Vert_F + \bigl\vert \tilde{\lambda}_i - {\lambda}_i \bigr\vert
\end{align*}
Therefore, given our approximation guarantee on the eigenvalues and that $\sqrt{\sum_{i=1}^{r'}(\widetilde{\lambda}_i - \lambda_i)^2} \le \epsilon$ implies $\max_{1\le i\le r'}\, \bigl\vert\widetilde{\lambda}_i - \lambda_i\bigr\vert \le \epsilon$, we can write the following
\begin{align}
\mathbb{P}\Bigl( \max_{1\le i\le r'} \bigl\Vert {K}_{\mathbf{X}|\rho\mathbf{Y}} \widetilde{\mathbf{v}}_i - {\lambda}_i\widetilde{\mathbf{v}}_i\bigr\Vert_2\, \ge\, 2\epsilon  \Bigr) \, \le \, \delta.
\end{align}
Note that the event under which we proved the closeness of the eigenvalues also requires that the Frobenius norm of $\Vert \widetilde{K}_{\mathbf{X}|\rho \mathbf{Y}}- {K}_{\mathbf{X}|\rho \mathbf{Y}} \Vert_F = \Vert \widetilde{K}_{\mathbf{X},\rho \mathbf{Y}}- {K}_{\mathbf{X},\rho \mathbf{Y}} \Vert_F\le \epsilon$ to hold. On the other hand, we note that ${K}_{\mathbf{X}|\rho\mathbf{Y}} = D{K}_{\mathbf{X},\rho\mathbf{Y}}= D\Phi^\top\Phi$ which leads to the conditional kernel covariance matrix $\Lambda_{\mathbf{X}|\rho\mathbf{Y}} = \Phi D\Phi^\top $. Therefore, if we define $\widetilde{\mathbf{u}}_i = \Phi \widetilde{\mathbf{v}}_i$ we will have
\begin{align*}
\bigl\Vert \Lambda_{\mathbf{X}|\rho\mathbf{Y}} \widetilde{\mathbf{u}}_i - \lambda_i \widetilde{\mathbf{u}}_i\bigr\Vert \, &=\, \bigl\Vert \Phi {K}_{\mathbf{X}|\rho\mathbf{Y}} \widetilde{\mathbf{v}}_i - {\lambda}_i\Phi\widetilde{\mathbf{v}}_i\bigr\Vert_2\\
\, &\le\, \bigl\Vert \Phi \bigr\Vert_2 \bigl\Vert {K}_{\mathbf{X}|\rho\mathbf{Y}} \widetilde{\mathbf{v}}_i - {\lambda}_i\widetilde{\mathbf{v}}_i\bigr\Vert_2
\\
\, &\le\, (1+\rho) \bigl\Vert {K}_{\mathbf{X}|\rho\mathbf{Y}} \widetilde{\mathbf{v}}_i - {\lambda}_i\widetilde{\mathbf{v}}_i\bigr\Vert_2
\end{align*}
We can combine the above results to show the following and complete the proof:
\begin{align}
\mathbb{P}\Bigl( \max_{1\le i\le r'} \bigl\Vert \Lambda_{\mathbf{X}|\rho\mathbf{Y}} \widetilde{\mathbf{u}}_i - {\lambda}_i\widetilde{\mathbf{u}}_i\bigr\Vert_2\, \ge\, 2(1+\rho)\epsilon  \Bigr) \, \le \, \delta.
\end{align}

\section{Additional Numerical Results}
\begin{figure*}[t]
    \centering    
    \includegraphics[width=0.99\textwidth]{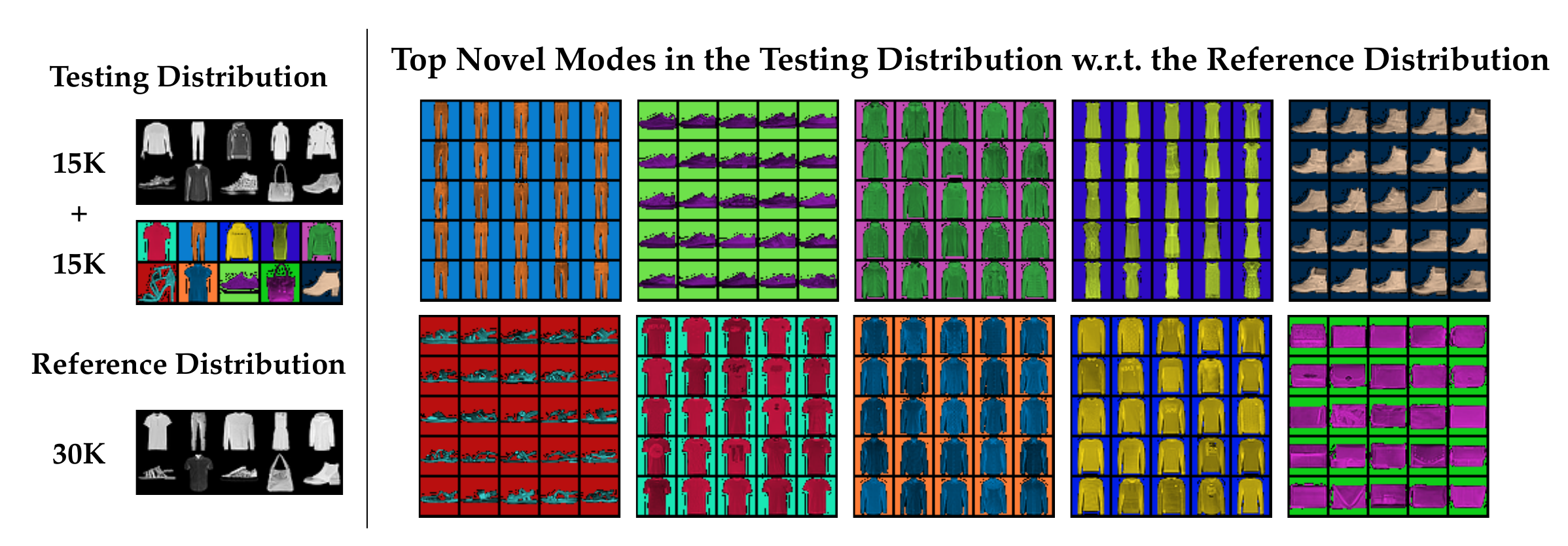}
    \caption{FINC-identified top 10 novel modes between Fashion-MNIST datasets. }\vspace{-4mm}
    \label{fig:color_fashion_mnist}  
\end{figure*}

\begin{figure}
    \centering    
    \includegraphics[width=0.99\textwidth]{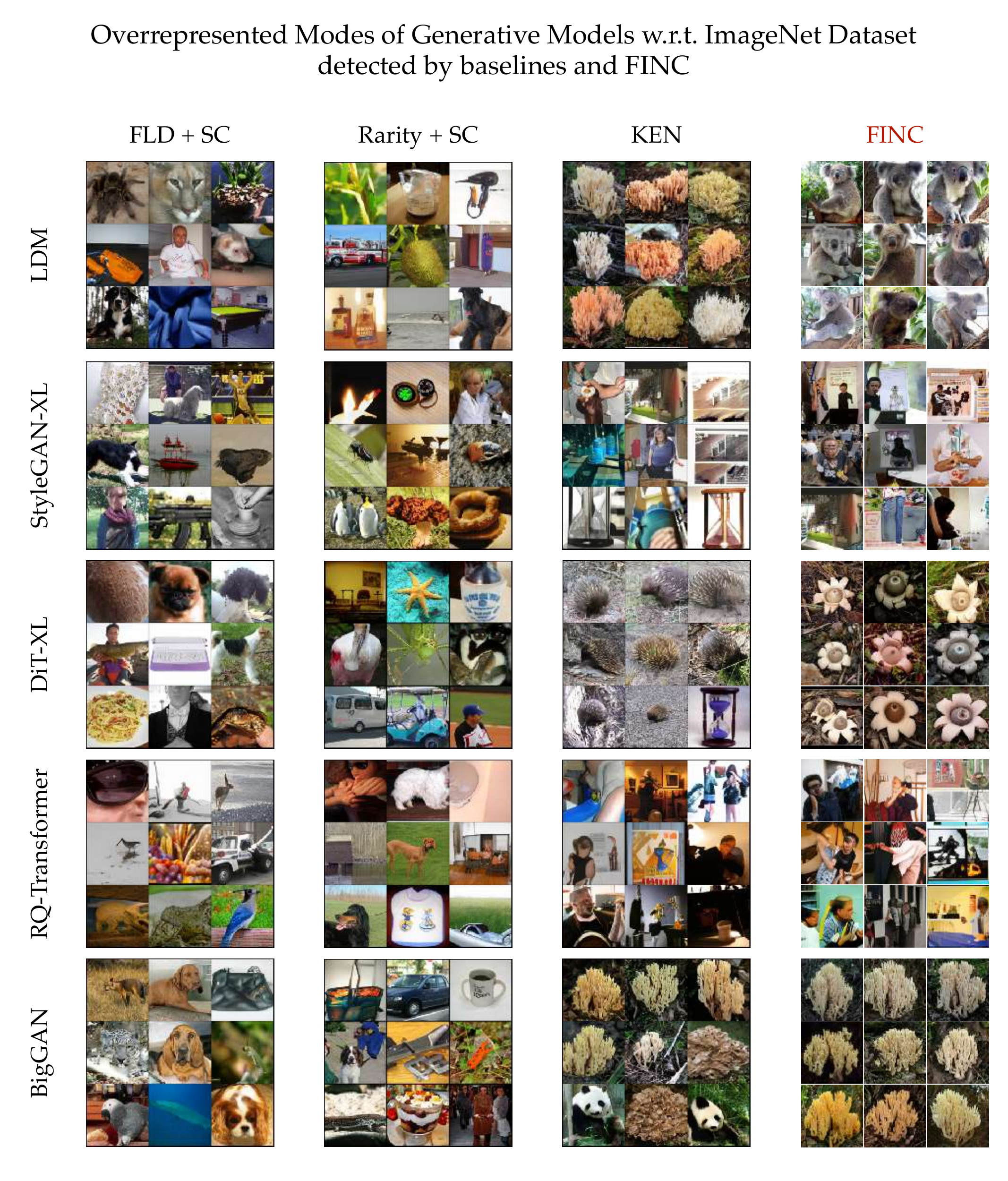}
    \caption{Baseline-identified and FINC-identified top overrepresented modes expressed by test generative models with a considerably higher frequency than in the reference ImageNet dataset. DINOv2 embedding is used. For baseline methods, we use the applicable largest sample size 10k in the GPU computation. For the FINC method, we use 50k samples instead.}
    \label{fig:supp_baseline_imgnet}  
\end{figure}

\begin{figure}
    \centering    
    \includegraphics[width=0.99\textwidth]{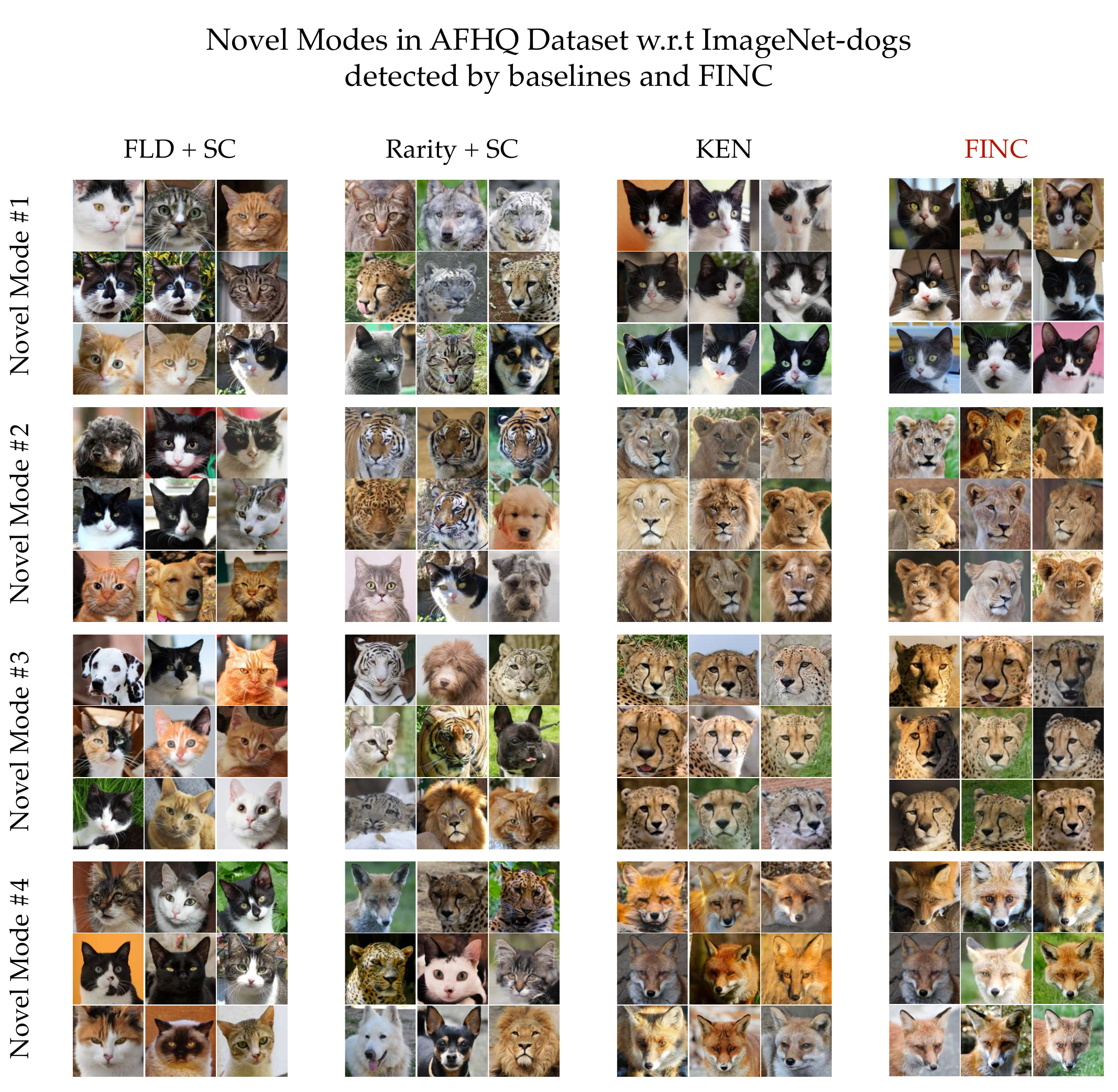}
    \caption{Baseline-identified and FINC-identified novel modes of AFHQ with respect to ImageNet-dog dataset. Inception-V3 embedding is used. For baseline methods, we use the applicable largest sample size 10k in the GPU computation. For the FINC method, we use 50k samples instead.}
    \label{fig:supp_baseline_afhq}  
\end{figure}

\begin{figure}[t]
    \centering    
    \includegraphics[width=0.99\textwidth]{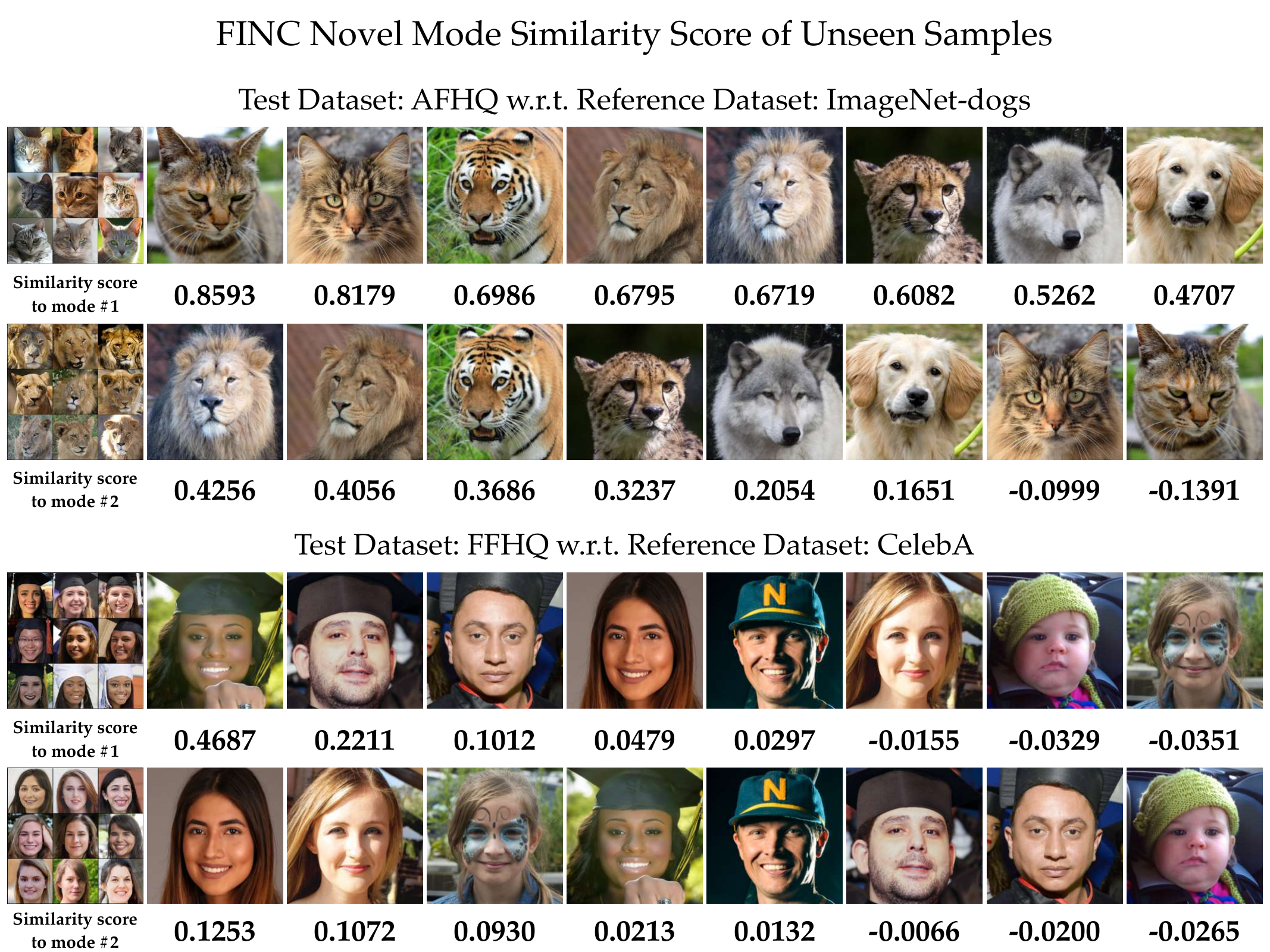}
    
    \caption{Computing FINC-assigned mode score for 8 unseen samples and 2 detected modes. Images looking more similar to the modes are supposed to have higher scores. (Top-half) AFHQ w.r.t. ImageNet-dogs, (Bottom-half) FFHQ w.r.t. CelebA.}
    \label{fig:score_func}  
\end{figure}

\begin{figure}
    \centering    \includegraphics[width=0.99\textwidth]{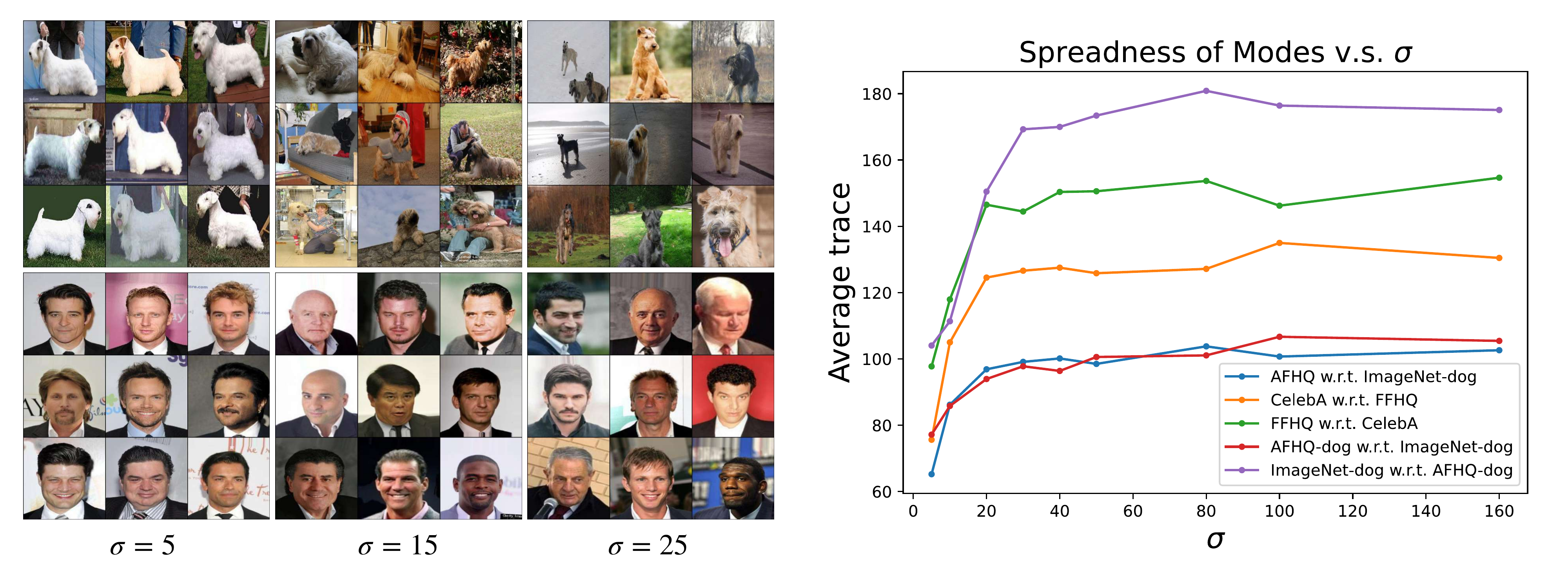}
    \caption{Effect analysis of hyperparameter $\sigma$. Modes captured by smaller $\sigma$ are more specific with similar samples and lower spreadness statistic, while modes captured by larger $\sigma$ are more general with diverse samples and higher spreadness statistic.}
    \label{fig:sigma_trend}  
\end{figure}

\begin{figure}
    \centering    \includegraphics[width=0.99\textwidth]{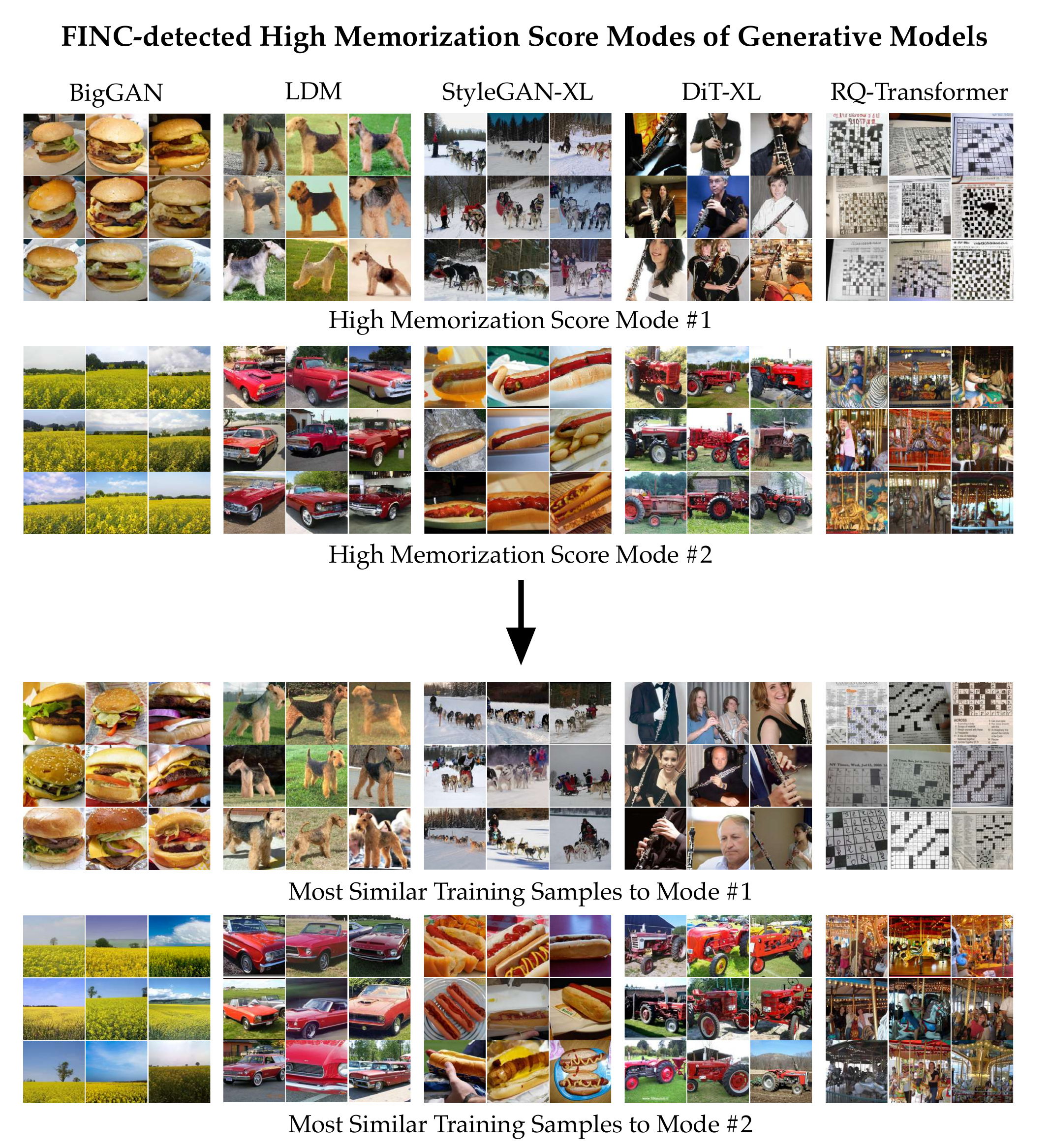}
    \caption{\textbf{Top: }The FINC-identified top-2 generated sample clusters with higher FLD memorization scores w.r.t. the ImageNet training dataset. The top 9 images with the maximum alignment with the detected mode’s eigenvectors are shown. \textbf{Bottom: }The most similar training samples retrieved from ImageNet training set.}
    \label{fig:memorized_dinov2_supp}  
\end{figure}

\subsection{Detailed Experiment Setting}

\textbf{Feature extraction:} Recent studies \cite{stein2023exposing,kyn2023} show current score-based evaluation may be biased in some applications, and the Inception-V3 embedding, which is retrieved from an ImageNet-pretrained network, may lead to unfair comparisons as well. To this end, these works propose replacing Inception-V3 with CLIP \cite{radford2021learning} and DINOv2 \cite{oquab2023dinov2} embeddings. Following \cite{stein2023exposing}, we adopt DINOv2 embedding in our image-based differential clustering experiments. Also, we downloaded the pre-trained generative models from DINOv2 \cite{stein2023exposing} repository. In our experiments, we present results both from Inception-V3 and DINOv2 embedding.

\textbf{Hyper-parameters selection and sample size:} Similar to the kernel-based evaluation in \cite{jalali2023information}, we chose the kernel bandwidth $\sigma^2$ by searching for the smallest $\sigma$ value resulting in a variance bounded by $0.01$. 
Also, we conducted the experiments using at least $m, n = 50$k sample sizes for the test and reference generative models, with the exception of dataset AFHQ and ImageNet-dogs, where the dataset's size was smaller than $50$k. For these datasets, we used the entire dataset in the differential clustering. For Fourier feature size $r$, we tested $r$ values in $\{1000,2000,4000\}$ to find the size obtaining an approximation error estimate below 1\%, and $r=2000$ consistently fulfilled the goal.

\subsection{Qualitative Comparison between baselines and FINC on ImageNet and AFHQ Dataset}

To compare the differential clustering results between the novelty score-based baseline methods and our proposed FINC method, we considered the baseline methods following the FLD score \cite{jiralerspong2023feature}, Rarity score \cite{han2023rarity}, and KEN score \cite{zhang2024interpretable}. In the experiments concerning the FLD and Rarity score-based baselines, we used the methods to detect novel samples, and subsequently we performed the spectral clustering from the TorchCluster package. For the baseline methods, we utilized the largest sample size 10k which did not lead to the memory overflow issue in our GPU processors. 

 Figure~\ref{fig:supp_baseline_imgnet} presents our numerical results on the ImageNet dataset, suggesting that the novel sample clusters in the Rarity score and FLD baselines may be semantically dissimilar in some cases. On the other hand, we observed that the novel samples within the clusters found by the KEN-based baseline method are more semantically similar. However, a few clusters found by the KEN method may contain more than one mode. However, the proposed FINC method seems to perform a semantically meaningful clustering of the modes in the large-scale ImageNet setting.

We also performed experiments on the AFHQ dataset containing only 15k images that is fewer than 1.4 million images in ImageNet-1K. The baseline methods FLD and KEN methods seemed to perform with higher accuracy on the AFHQ dataset in comparison to the ImageNet-1K dataset. As displayed in Figure~\ref{fig:supp_baseline_afhq}, FLD seems capable of detecting cat-like modes. The proposed FINC and the baseline KEN method led to nearly identical found modes.

\subsection{Experimental Results on Video Datasets}
\begin{figure}[h]
    \centering
    \begin{subfigure}[t]{0.32\textwidth}
        \centering
        \caption*{Novel Mode \#1}
        \includegraphics[width=\linewidth]{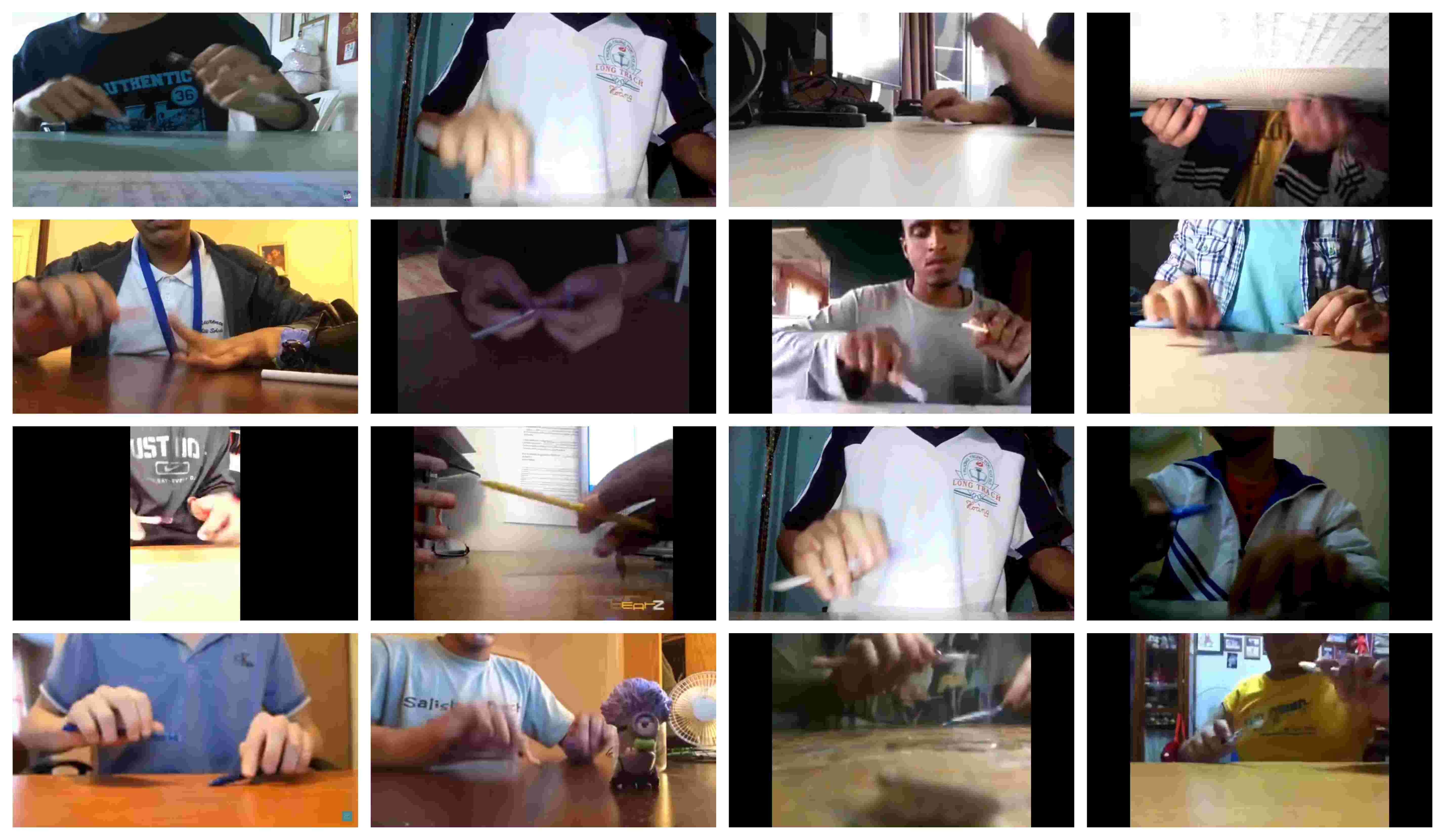}
    \end{subfigure}
    \begin{subfigure}[t]{0.32\textwidth}
        \centering
        \caption*{Novel Mode \#2}
        \includegraphics[width=\linewidth]{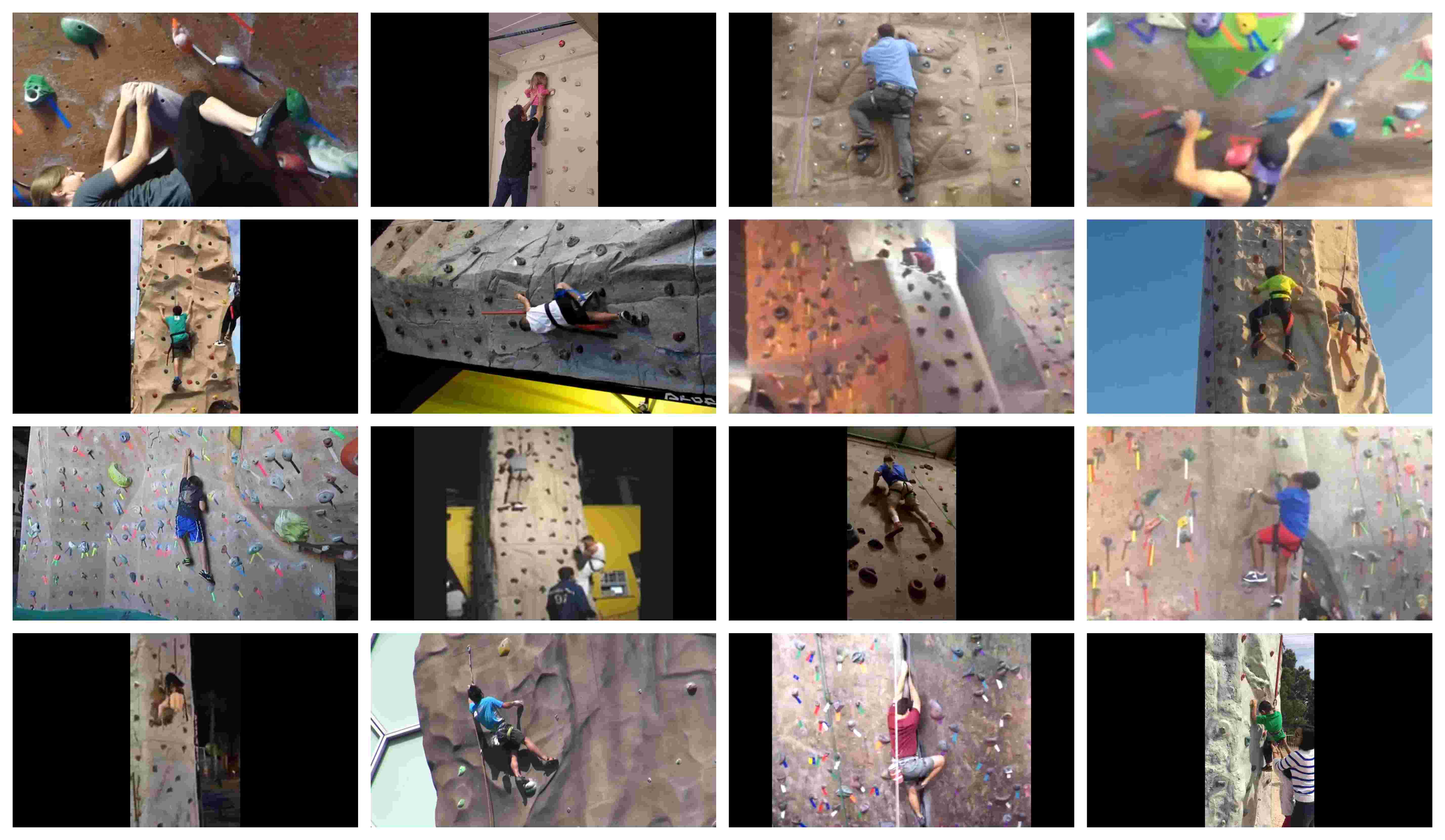}
    \end{subfigure}
    \begin{subfigure}[t]{0.32\textwidth}
        \centering
        \caption*{Novel Mode \#3}
        \includegraphics[width=\linewidth]{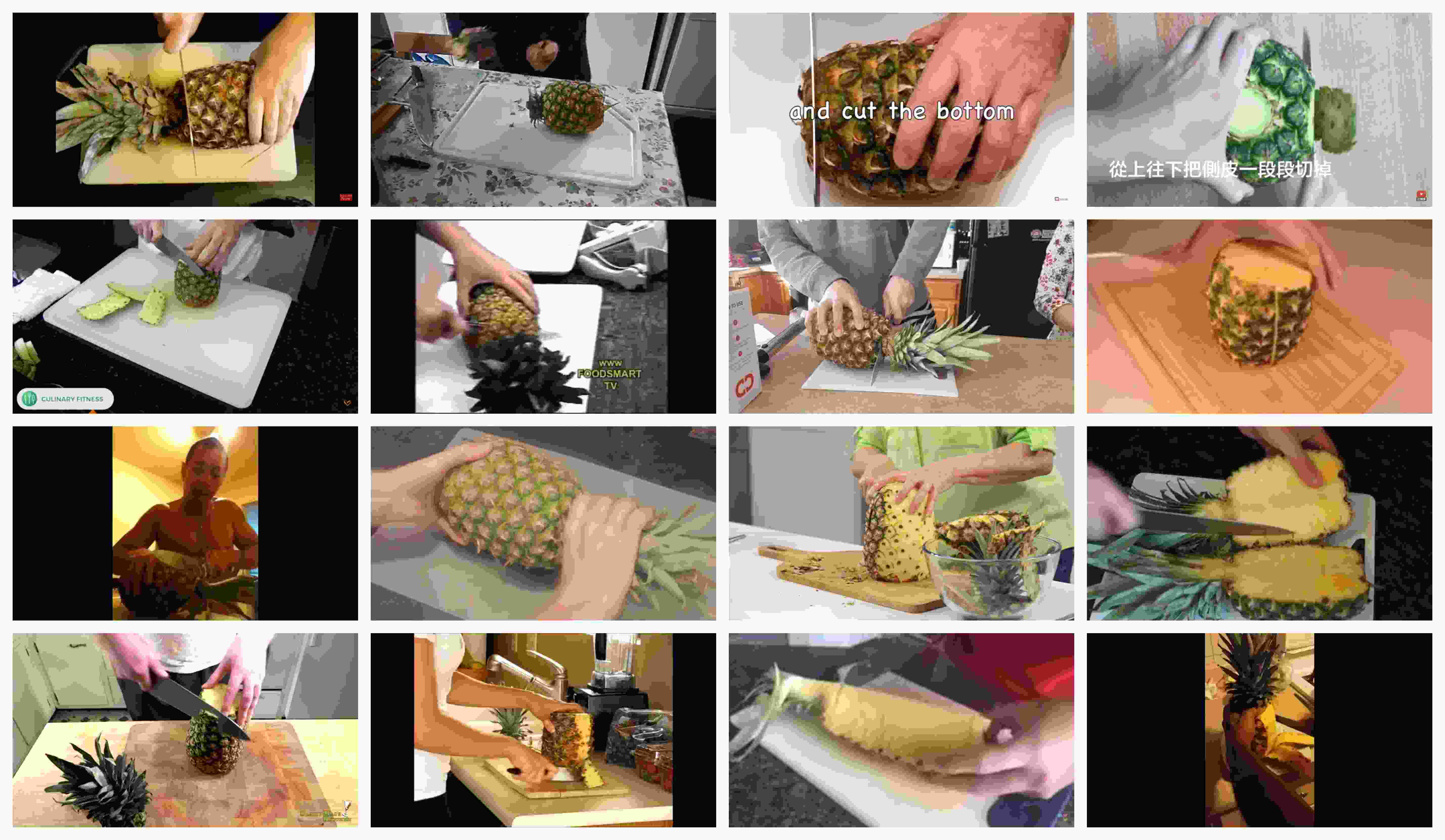}
    \end{subfigure}\\
    \begin{subfigure}[t]{0.32\textwidth}
        \centering
        \caption*{Novel Mode \#4}
        \includegraphics[width=\linewidth]{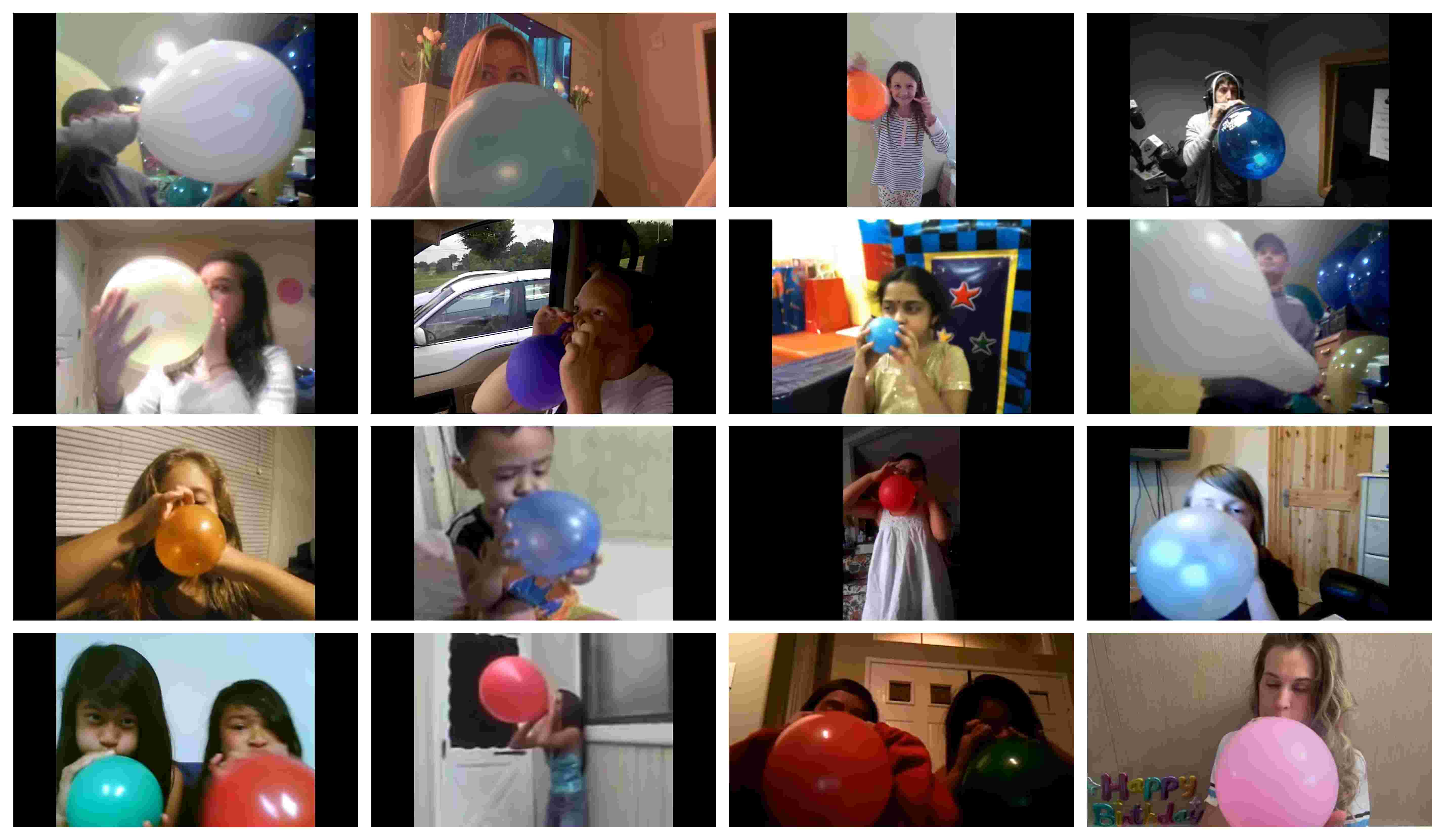}
    \end{subfigure}
        \begin{subfigure}[t]{0.32\textwidth}
        \centering
        \caption*{Novel Mode \#5}
        \includegraphics[width=\linewidth]{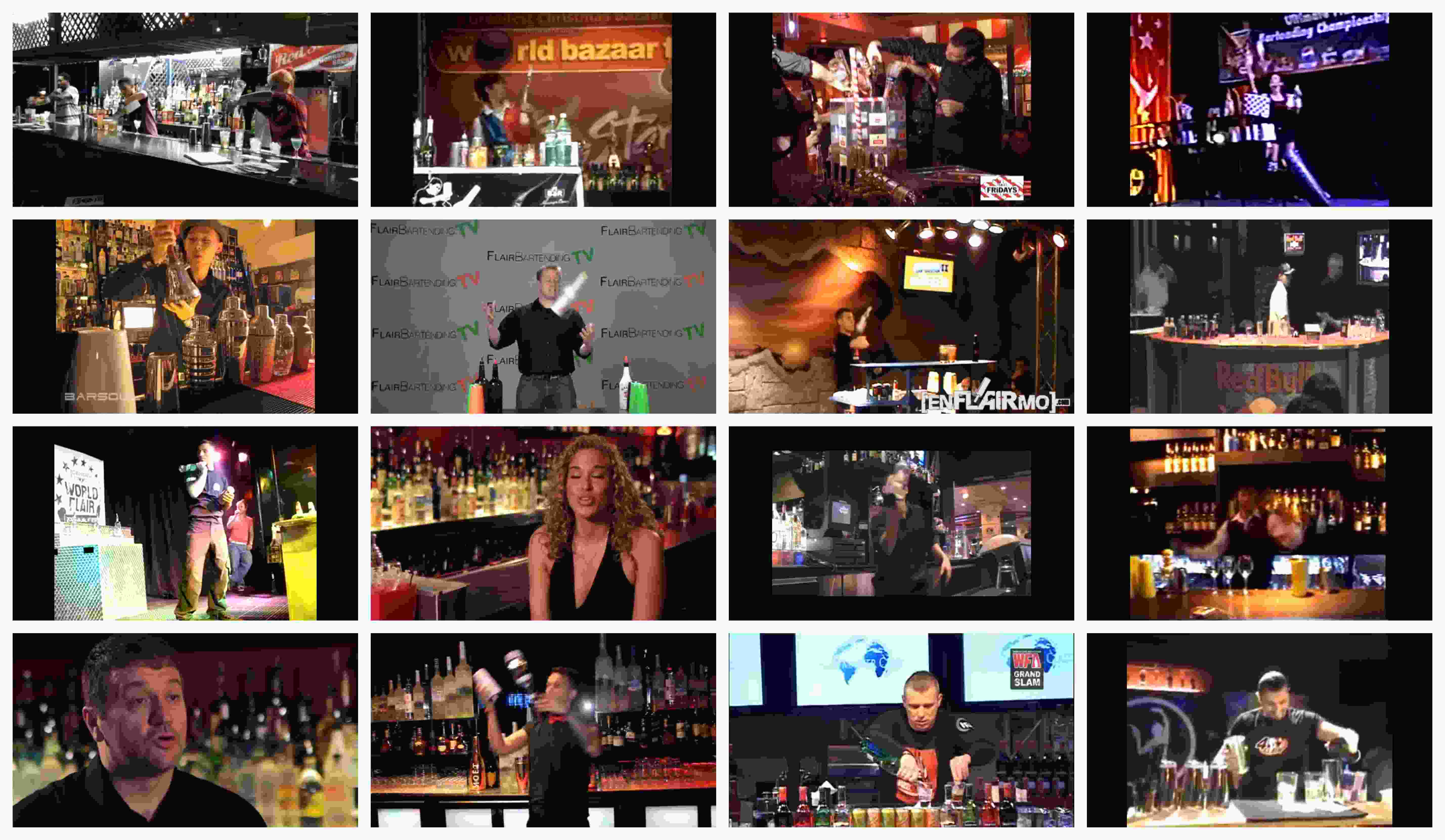}
    \end{subfigure}
    \begin{subfigure}[t]{0.32\textwidth}
        \centering
        \caption*{Novel Mode \#6}
        \includegraphics[width=\linewidth]{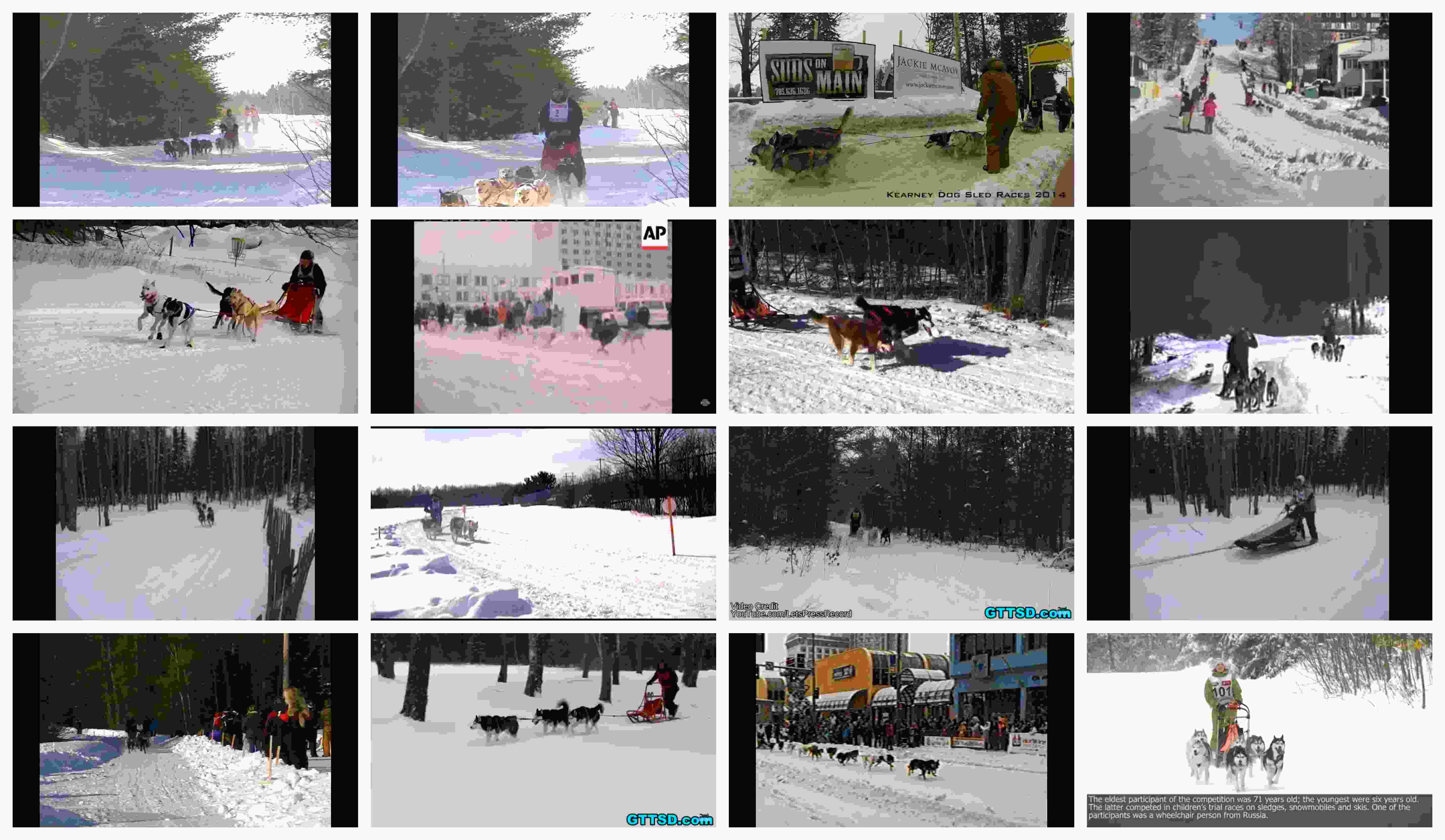}
    \end{subfigure}\\
    \begin{subfigure}[t]{0.32\textwidth}
        \centering
        \caption*{Novel Mode \#7}
        \includegraphics[width=\linewidth]{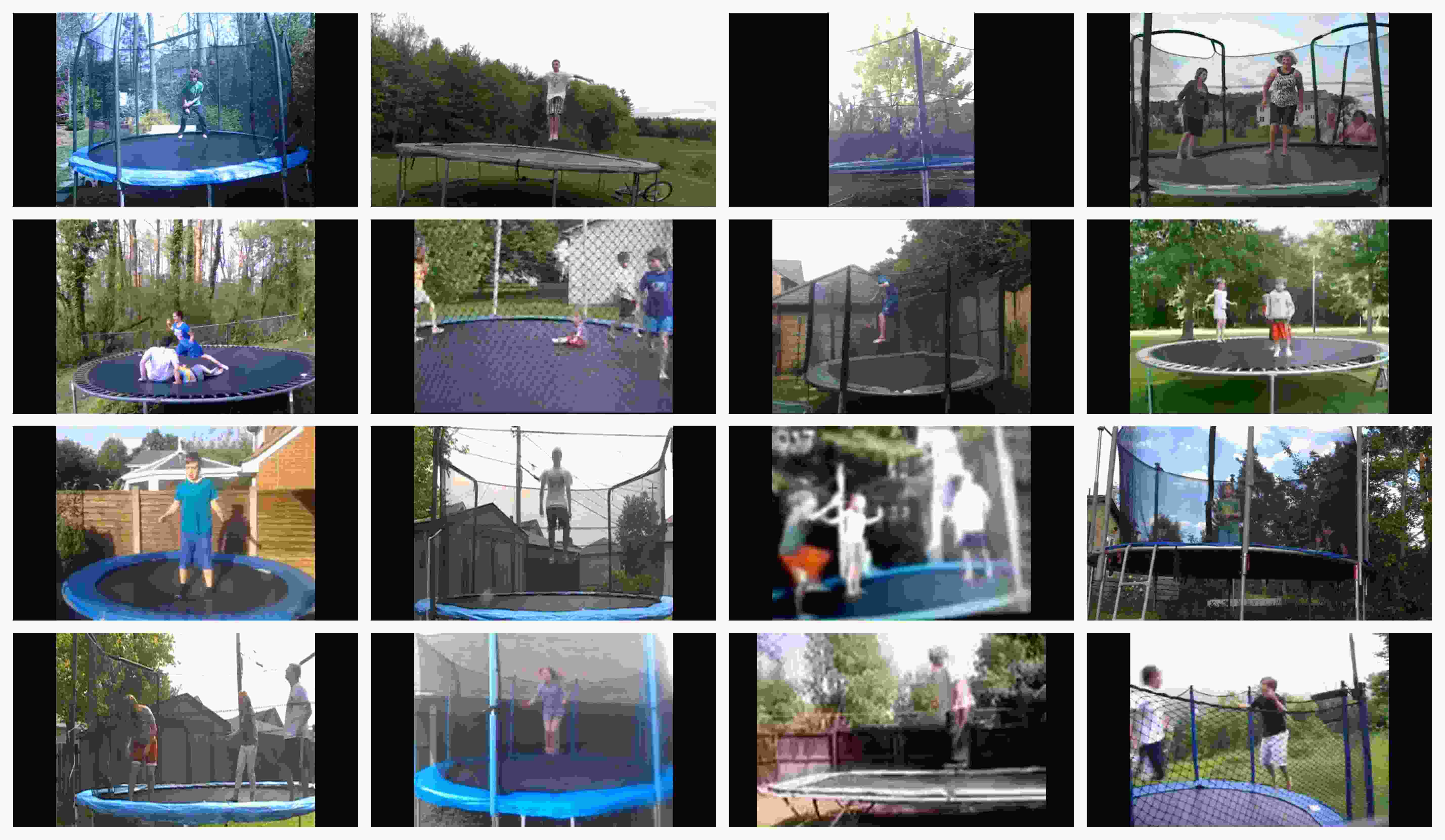}
    \end{subfigure}
    \begin{subfigure}[t]{0.32\textwidth}
        \centering
        \caption*{Novel Mode \#8}
        \includegraphics[width=\linewidth]{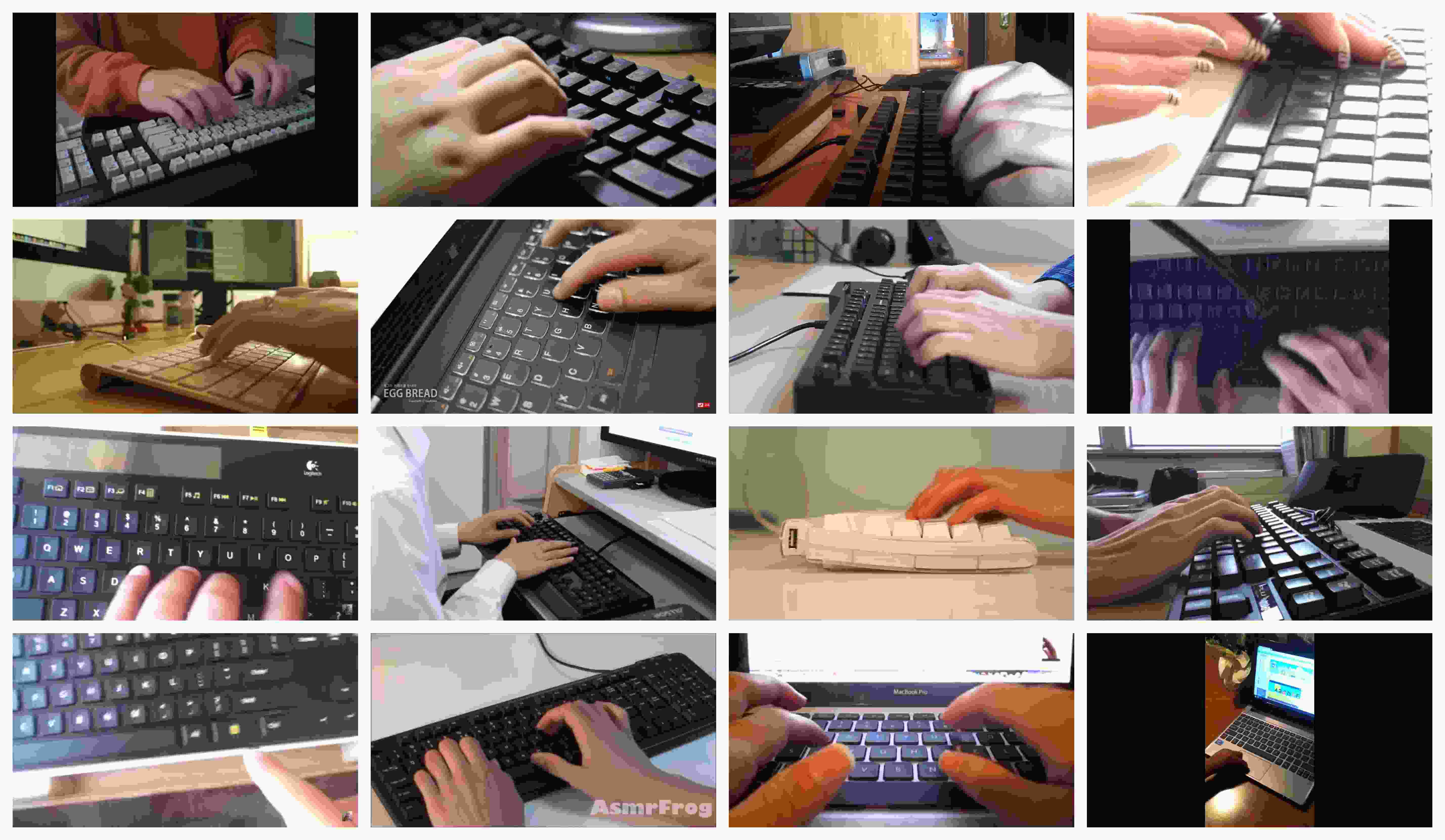}
    \end{subfigure}
        \begin{subfigure}[t]{0.32\textwidth}
        \centering
        \caption*{Novel Mode \#9}
        \includegraphics[width=\linewidth]{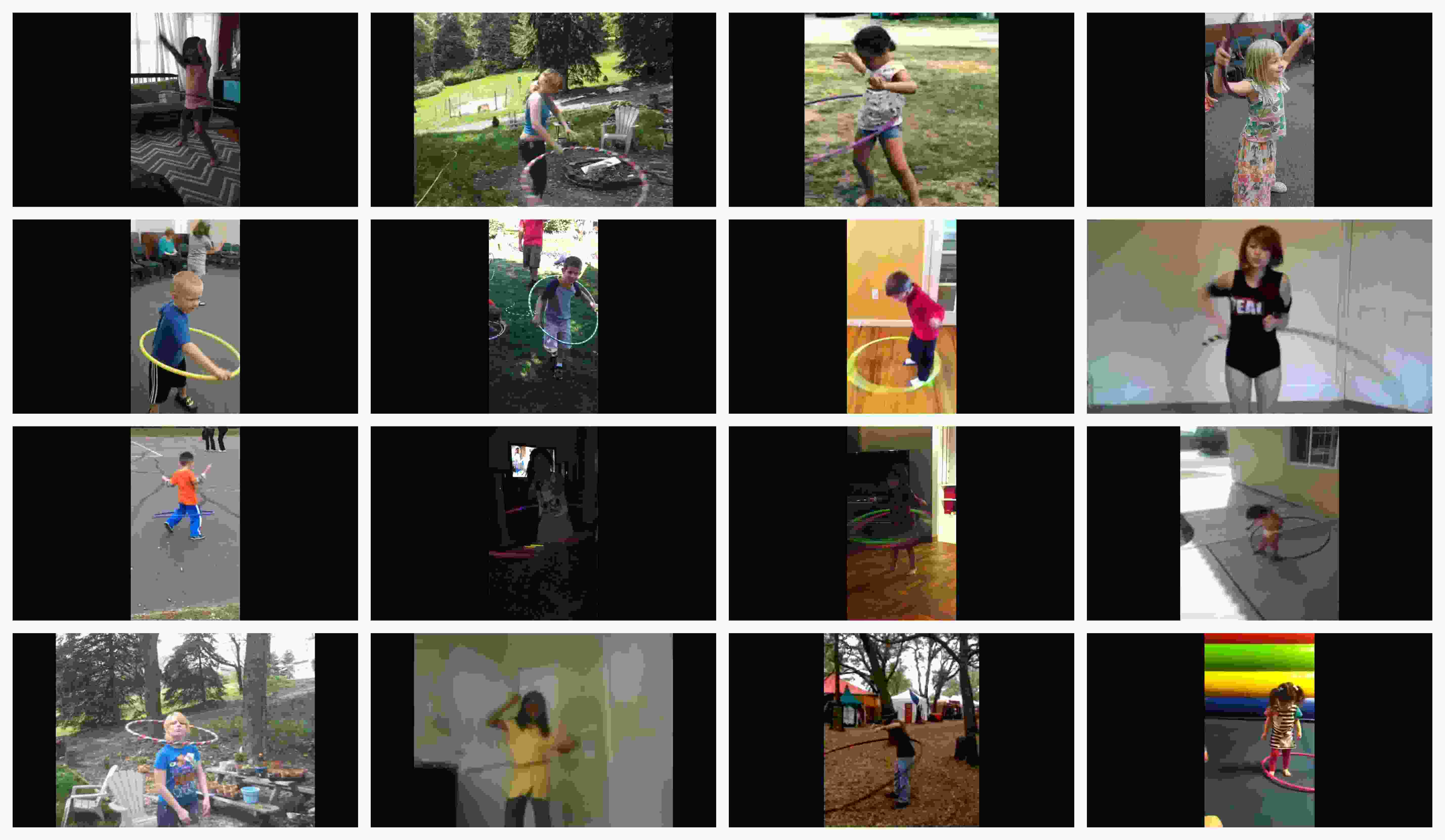}
    \end{subfigure}
    \caption{Top 9 novel modes of Kinetics-400 dataset based on FINC evaluation with UCF101 as the reference dataset.}
  \label{fig:kinetics}
\end{figure}

In this section, we compare Kinetics-400 \cite{kay2017kinetics} with UCF101 \cite{soomro2012ucf101} dataset.
Kinetics-400 includes 400 human action categories, each with a minimum of 400 video clips depicting the action, while UCF101 consists of 101 actions.
Similar to video evaluation metrics \cite{Saito2020, unterthiner2019accurate} we used the I3D pre-trained model \cite{Carreira_i3d} which maps each video to a 1024-vector feature. In this experiment, we considered Kinetics-400 as the test and UCF101 as our reference dataset. The top 9 novel modes are shown in Figure~\ref{fig:kinetics} and none of these categories are included in UCF101.

\subsection{Experimental Results on Text-to-Image Models}

\subsubsection{Evaluation of text-to-image Models}
\begin{figure}
    \begin{subfigure}[t]{0.23\textwidth}
        \centering
        \caption*{Novel Mode \#1}
        \includegraphics[width=\linewidth]{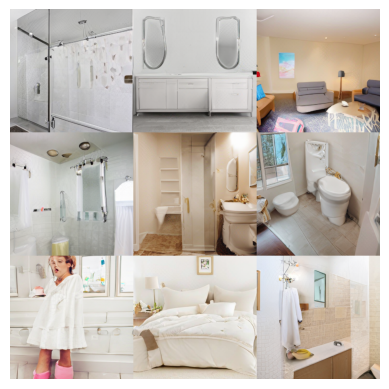}
    \end{subfigure}
    \begin{subfigure}[t]{0.23\textwidth}
        \centering
        \caption*{Novel Mode \#2}
        \includegraphics[width=\linewidth]{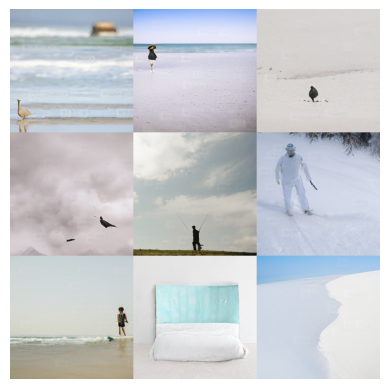}
    \end{subfigure}
    \begin{subfigure}[t]{0.23\textwidth}
        \centering
        \caption*{Less-frequent Mode \#1}
        \includegraphics[width=\linewidth]{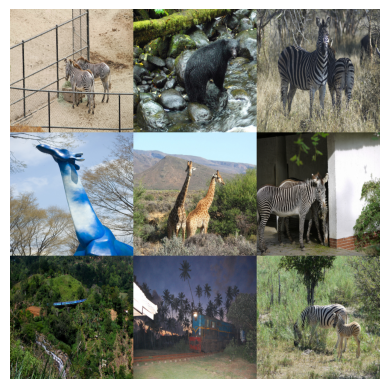}
    \end{subfigure}
    \begin{subfigure}[t]{0.23\textwidth}
        \centering
        \caption*{Less-frequent Mode \#2}
        \includegraphics[width=\linewidth]{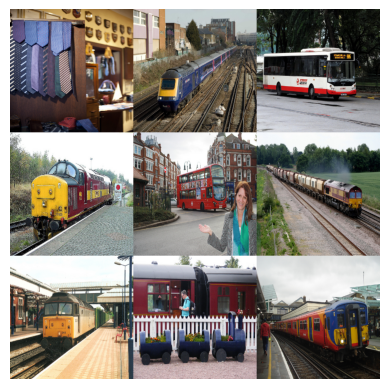}
    \end{subfigure}
    \centering
    \caption{Identified top-2 novel modes and top-2 less-frequent modes on the GigaGAN text-to-image model generated on the MS-COCO dataset.}
    \label{fig:gigagan}
\end{figure}

We assessed GigaGAN \cite{kang2023gigagan}, a state-of-the-art text-to-image generative model on the MS-COCO dataset \cite{lin2015microsoft}. We conducted a zero-shot evaluation on the 30K images from the MS-COCO validation set. As shown in Figure~\ref{fig:gigagan}, the model showed some novelty in generating pictures of a bathroom and sea with a bird or human on it. However, our FINC-based evaluation suggests that the model may be less capable in producing images of zebras, giraffes, or trains.

\begin{figure}[t]
    \centering    \includegraphics[width=0.85\linewidth]{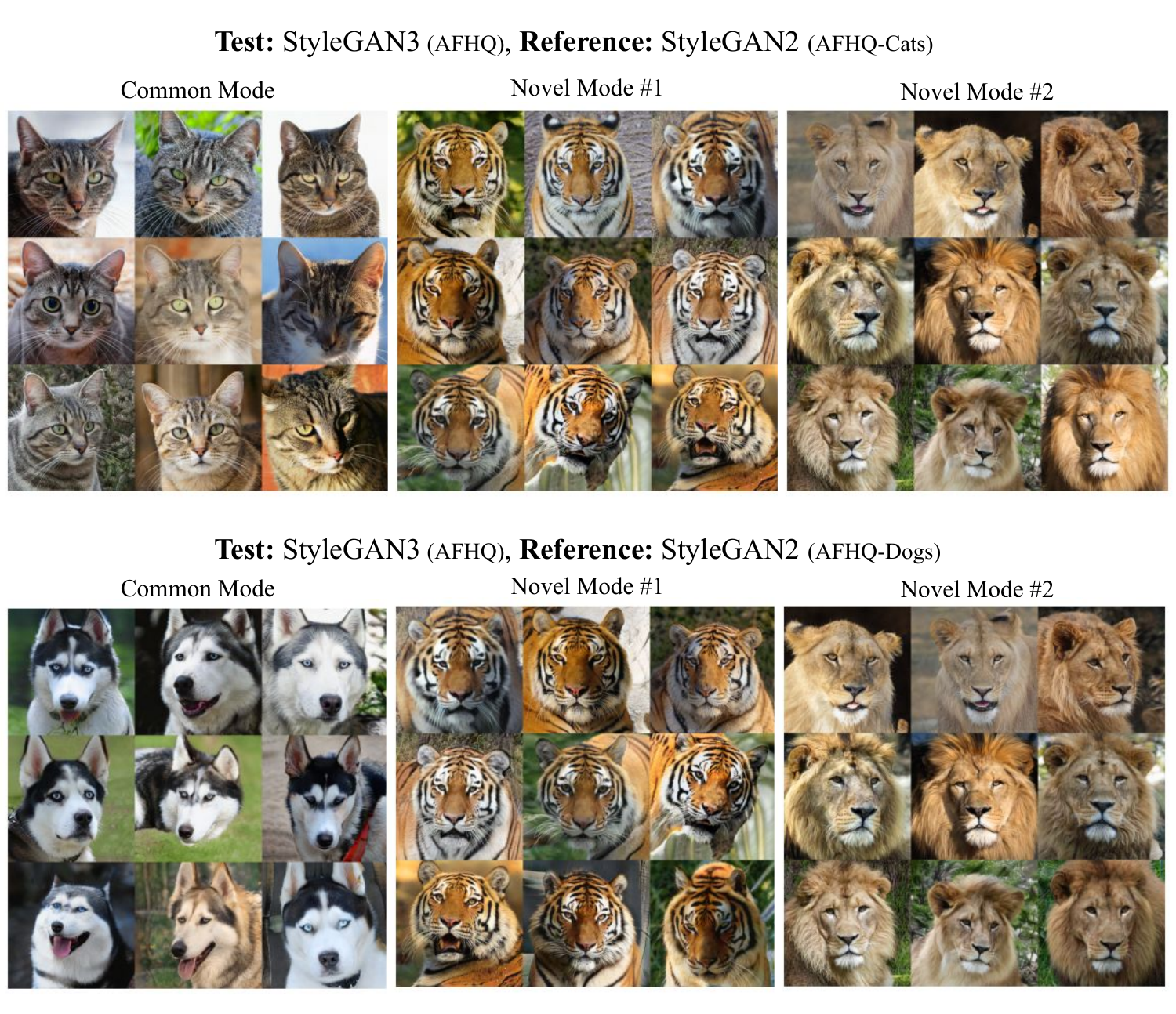}
    \caption{Identified top-2 novel modes and the most common mode between the StyleGAN3 and StyleGAN2 trained on AFHQ}
    \label{fig:afhq}
\end{figure}

\subsubsection{Effect of different feature extractors}
\begin{figure}
    \includegraphics[width=0.91\linewidth]{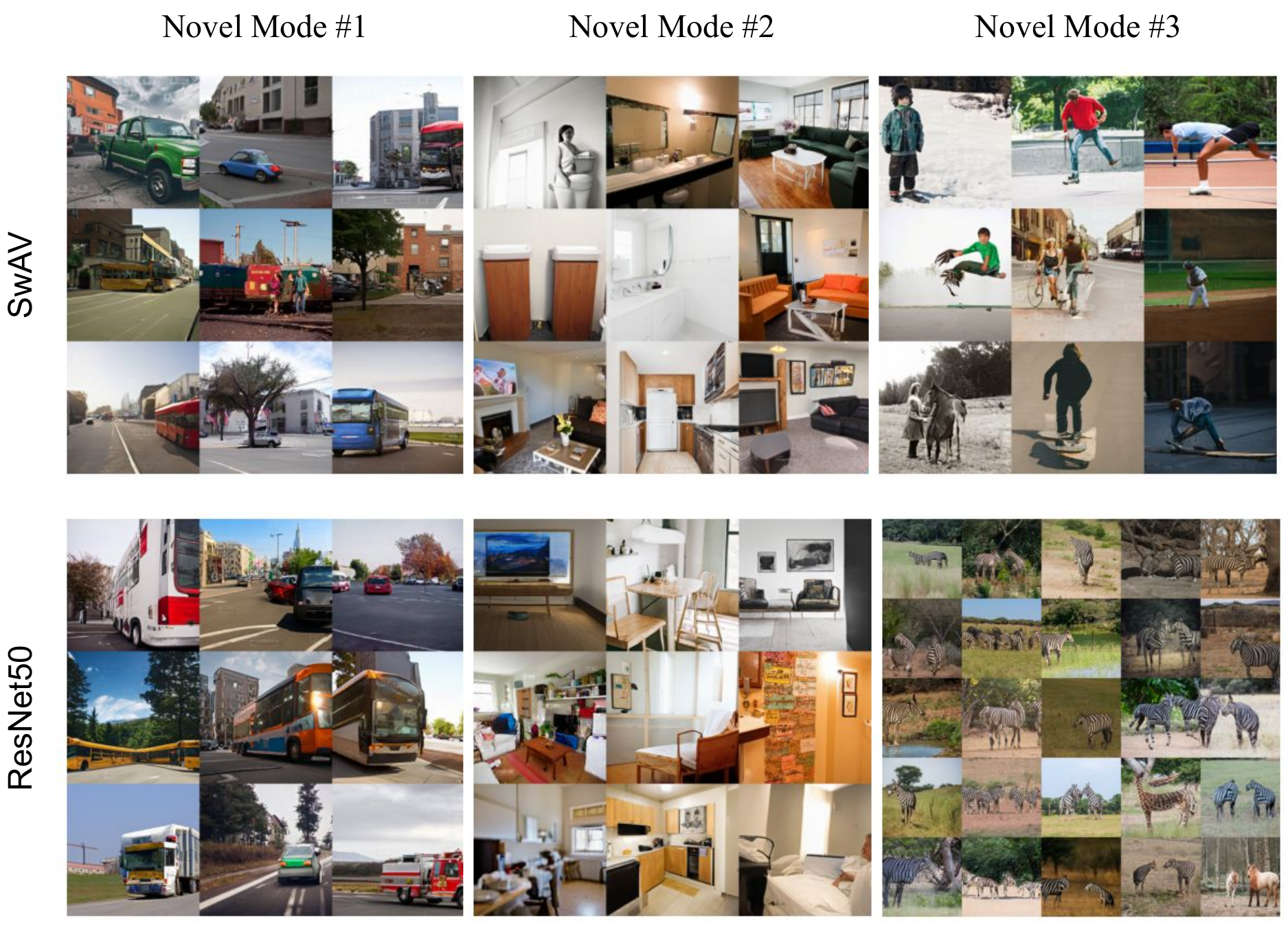}
    \centering
    \caption{Identified top-3 novel modes on the GigaGAN text-to-image model generated on the MS-COCO dataset using SwAV and ResNet50 pre-trained models as feature extractors.}
    \label{fig:effect_feature_extractors}
\end{figure}

As shown in the literature \cite{kyn2023}, Inception-V3 may lead to a biased evaluation metric. In this experiment, instead of using pre-trained Inception-V3 to extract features, we used pre-trained models SwAV \cite{caron2021unsupervised} and ResNet50 \cite{he2015deep} as feature extractors and attempted to investigate the effect of different pre-trained models on our proposed method. 
Figure~\ref{fig:effect_feature_extractors} shows top-3 novel models of GigaGAN \cite{kang2023gigagan} using SwAV and ResNet50 models. The top two novel modes were the same among the different feature extractors indicating that our method is not biased on the pre-trained networks.
In the third mode, SwAV and ResNet50 results were different showing that each of these models captured different novel modes based on the features they extracted from the data.

\subsection{Detection of Sample Clusters with Higher and Lower CLIPScores}
\begin{figure*}
    \centering    
    \includegraphics[width=\textwidth]{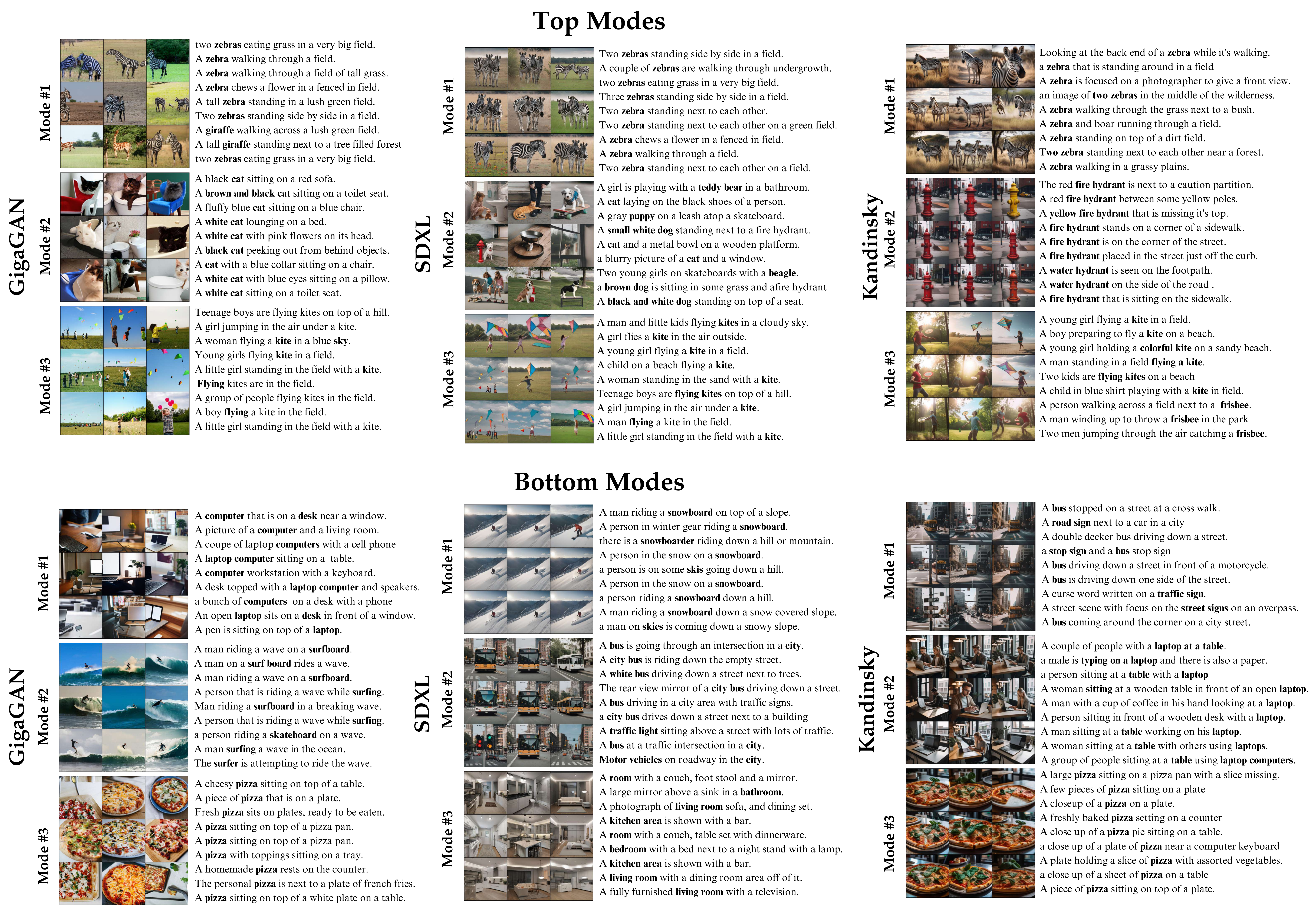}
    \vspace{-2mm}
    \caption{The FINC-identified top-3 and bottom-3 sample clusters according to the CLIPScore of text-to-image generative models's data.}\vspace{-4mm}
    \label{fig:clip_top_bottom}  
\end{figure*}

The CLIPScore \cite{hessel2021clipscore} is a popular sample-based metric to measure the alignment of the generated image by a text-to-image model. To identify the sample types with the highest and lowest CLIPScores, we ran a differential clustering experiment between the CLIPScore-based top 20\% and the bottom 20\% of samples generated by three text-to-image models, GigaGAN, SD-XL, and Kandinsky in response to text prompts from MS-COCO dataset. In Figure~\ref{fig:clip_top_bottom}, we show the top-3 and bottom-3 FINC-identified modes with the maximum and minimum CLIPScores for each text-to-image model. We observe similar clusters in the top modes, indicating that CLIPScore performs better in clusters related to zebras or cats/dogs, however, it does not perform well in bottom clusters such as buses or electronic devices.

\subsection{Detection of High Memorization-Score Modes}

In the literature,  memorization scores have been proposed to detect memorizing training samples by generative models. Although sample-level memorization metrics have been used to quantify and rank the similarity of every generated sample to training data, these scores do not directly reveal which sample types could be memorized more frequently by a generative model. To address this task, we apply FINC to conduct differential clustering on the groups of samples with low and high memorization FLD scores.

To do this, we evaluated and ranked the sample-level memorization score for all the generated samples using the FLD memorization metric with respect to the ImageNet training samples. Subsequently, we separated the two groups f samples with the top 20\% and bottom 20\% of FLD memorization scores. 
We performed differential clustering via FINC between the two groups, and used the eigenvalues/eigenvectors computed by Algorithm~\ref{algo:KEN} to quantify and rank mode-level memorization scores. Figure~\ref{fig:memorized_dinov2_supp} displays the top two detected sample types that occur more frequently in the high-memorization samples of the generative models, and the most similar ImageNet training samples to the generated modes.

\subsection{Comparative Mode Frequency Analysis between Generative Models}

\subsubsection{Evaluation of StyleGAN-XL on ImageNet}
\begin{figure}[t]
    \centering    
    \includegraphics[width=0.99\textwidth]{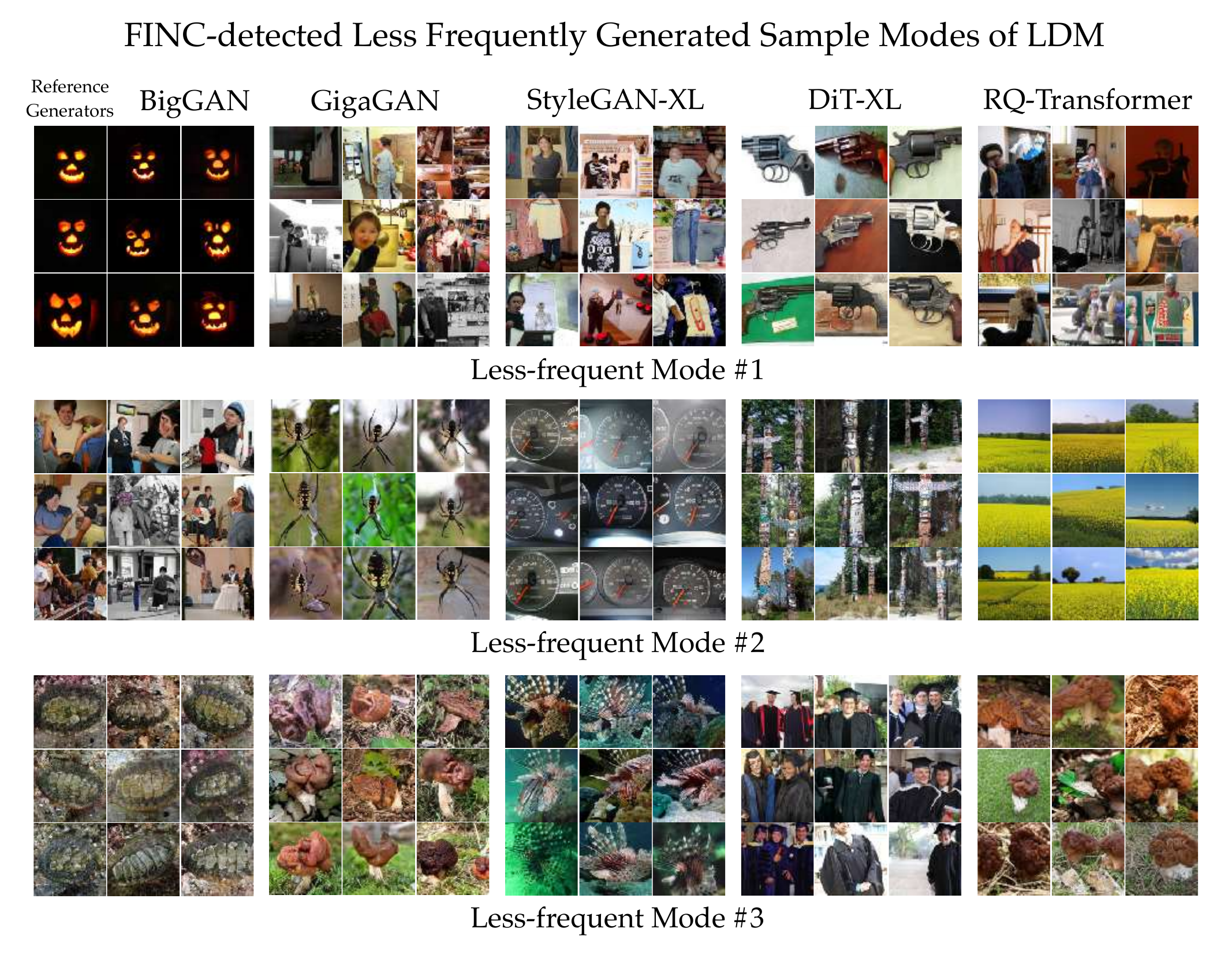}
    \caption{Representative samples from FINC-detected top-3 modes with the maximum frequency gap between LDM (lower frequency) and reference generative models. Embedding is from DINOv2.}
    \label{fig:lower_freq_ldm_dinov2}  
\end{figure}

\begin{figure}
    \centering    
    \includegraphics[width=0.99\textwidth]{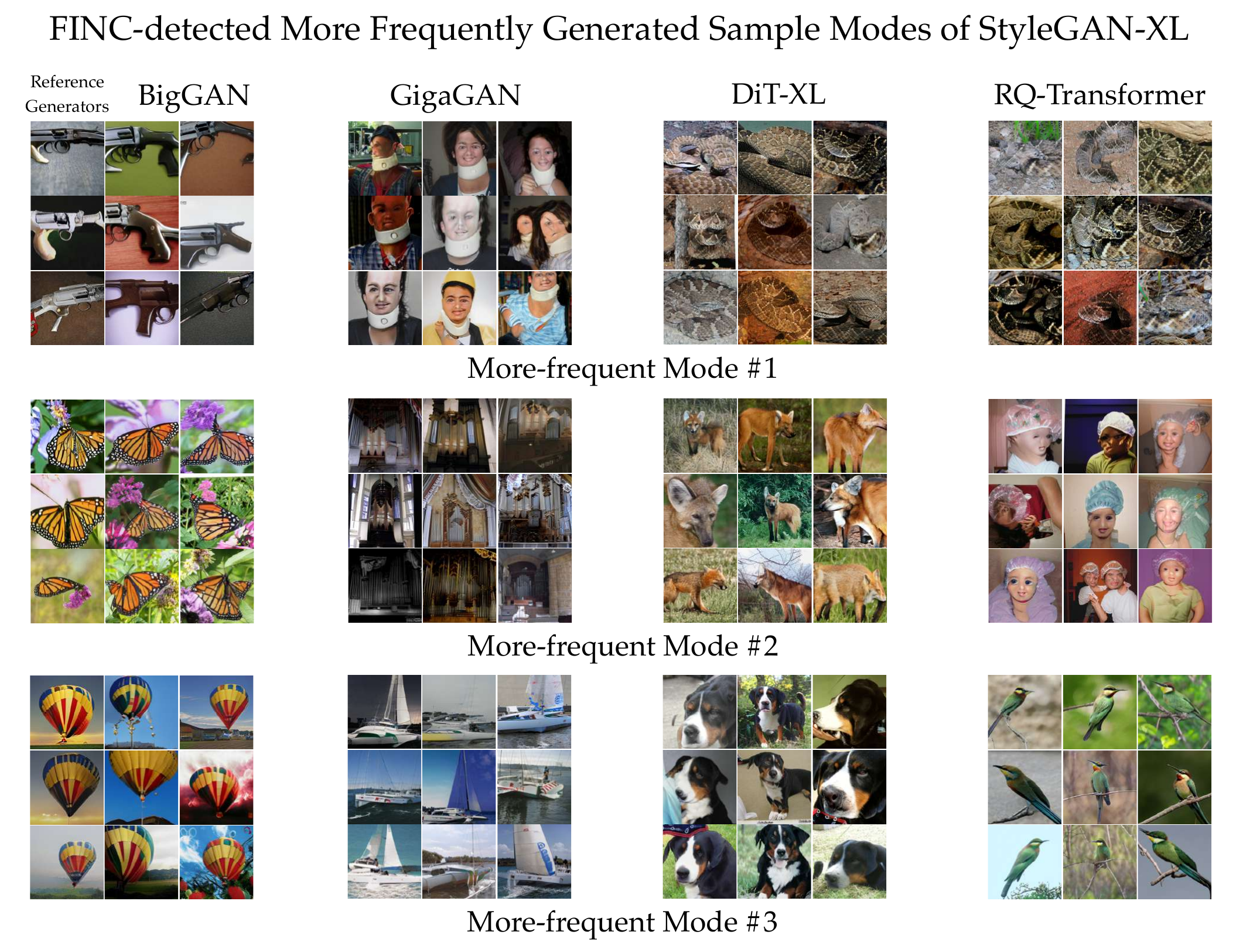}
    \caption{Representative samples from FINC-detected top-3 modes with the maximum frequency gap between StyleGAN-XL (higher frequency) and reference generative models.}
    \label{fig:higher_freq_styxl}  
\end{figure}

\begin{figure}
    \centering    
    \includegraphics[width=0.99\textwidth]{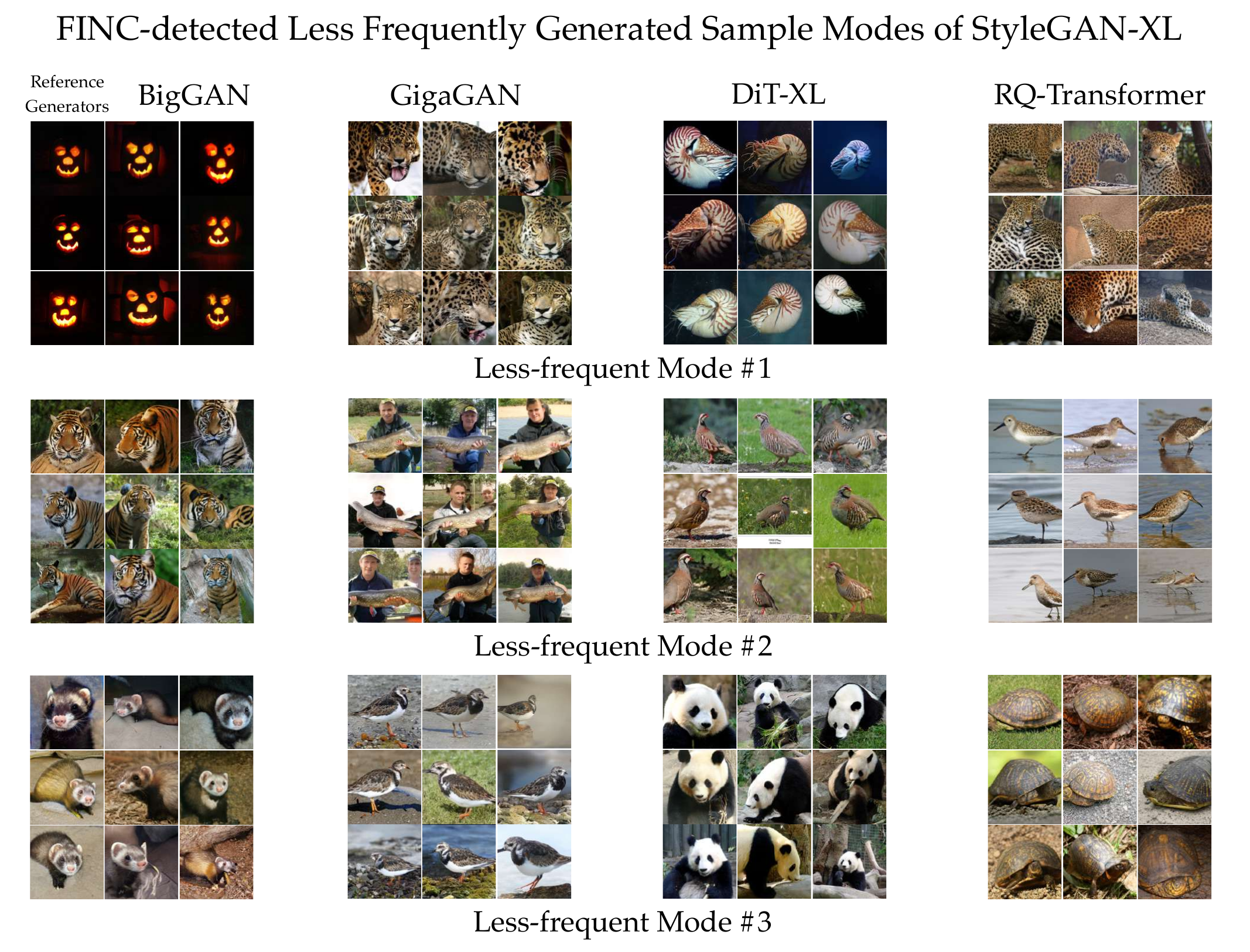}
    \caption{Representative samples from FINC-detected top-3 modes with the maximum frequency gap between StyleGAN-XL (lower frequency) and reference generative models.}
    \label{fig:lower_freq_styxl}  
\end{figure}

Figures \ref{fig:higher_freq_styxl} and \ref{fig:lower_freq_styxl} illustrate the more frequently and less frequently generated modes of StyleGAN-XL in comparison to the reference models. These figures repeat the experiments presented in Figures \ref{fig:higher_freq_ldm_dinov2} and \ref{fig:lower_freq_ldm_dinov2} on StyleGAN-XL.

\subsubsection{Novel Modes of DiT-XL on ImageNet} 
\begin{figure}[t]
    \centering    
    \includegraphics[width=0.99\textwidth]{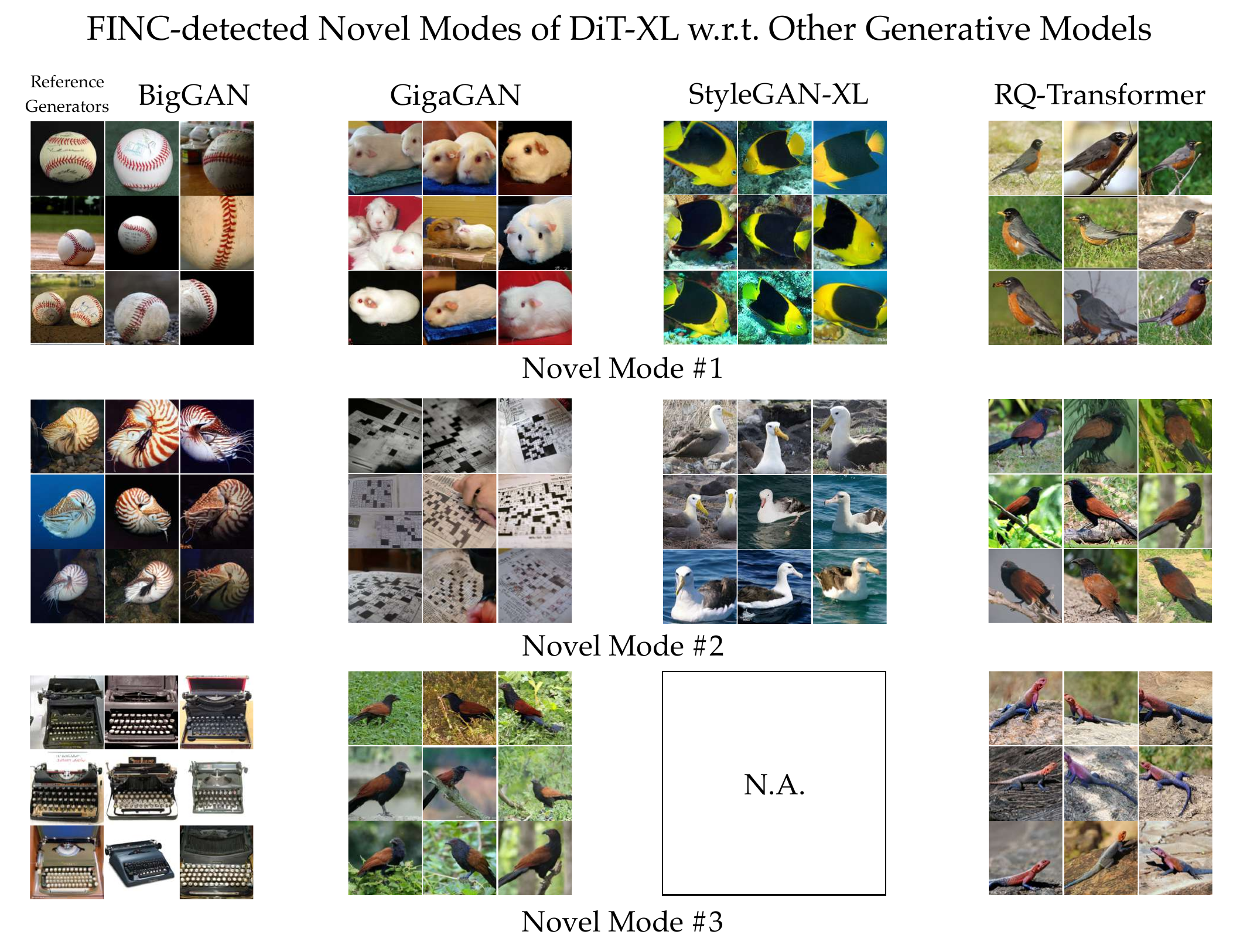}
    \caption{Representative samples from FINC-detected top-3 novel modes (novelty threshold $\rho=5$) of DiT-XL compared to reference generative models. "N.A." indicates that DiT-XL has fewer than three modes with a frequency $\rho=5$ times higher than the reference model. Additionally, it is worth noting that FINC does not detect any modes of DiT-XL with a frequency $\rho=5$ times higher than that of LDM.}
    \label{fig:novel_dit_imgnet}  
\end{figure}

Figure \ref{fig:novel_dit_imgnet} illustrates the novel modes of DiT-XL with novelty threshold $\rho=5$ in comparison to the reference models. For this experiment, we use a sample size of 100k in each distribution.

\subsection{Analyzing the Effect of $\sigma$ Hyperparameter}

We examined the effect of  $\sigma$ hyper-parameter of the Gaussian kernel in \eqref{Eq: Gaussian kernel}. We used the trace of the covariance matrix to measure the heterogeneity of samples within each mode. Based on our numerical evaluations shown in Figure~\ref{fig:sigma_trend}, we observed that smaller values of $\sigma$ could capture more specific modes containing samples with a seemingly lower variety. On the other hand, a greater $\sigma$ value seemed to better capture more general modes, displaying diverse samples and a higher variance statistic.

\subsection{Computational Software and Hardware}
We leverage the PyTorch framework for doing the experiments. All experiments are conducted in a cluster configured with  dual 24-core CPU processors with 500 GB memory and 8 RTX 3090 GPU processors with 24 GB memory.

\begin{figure}[t]
    \centering    
    \includegraphics[width=0.99\textwidth]{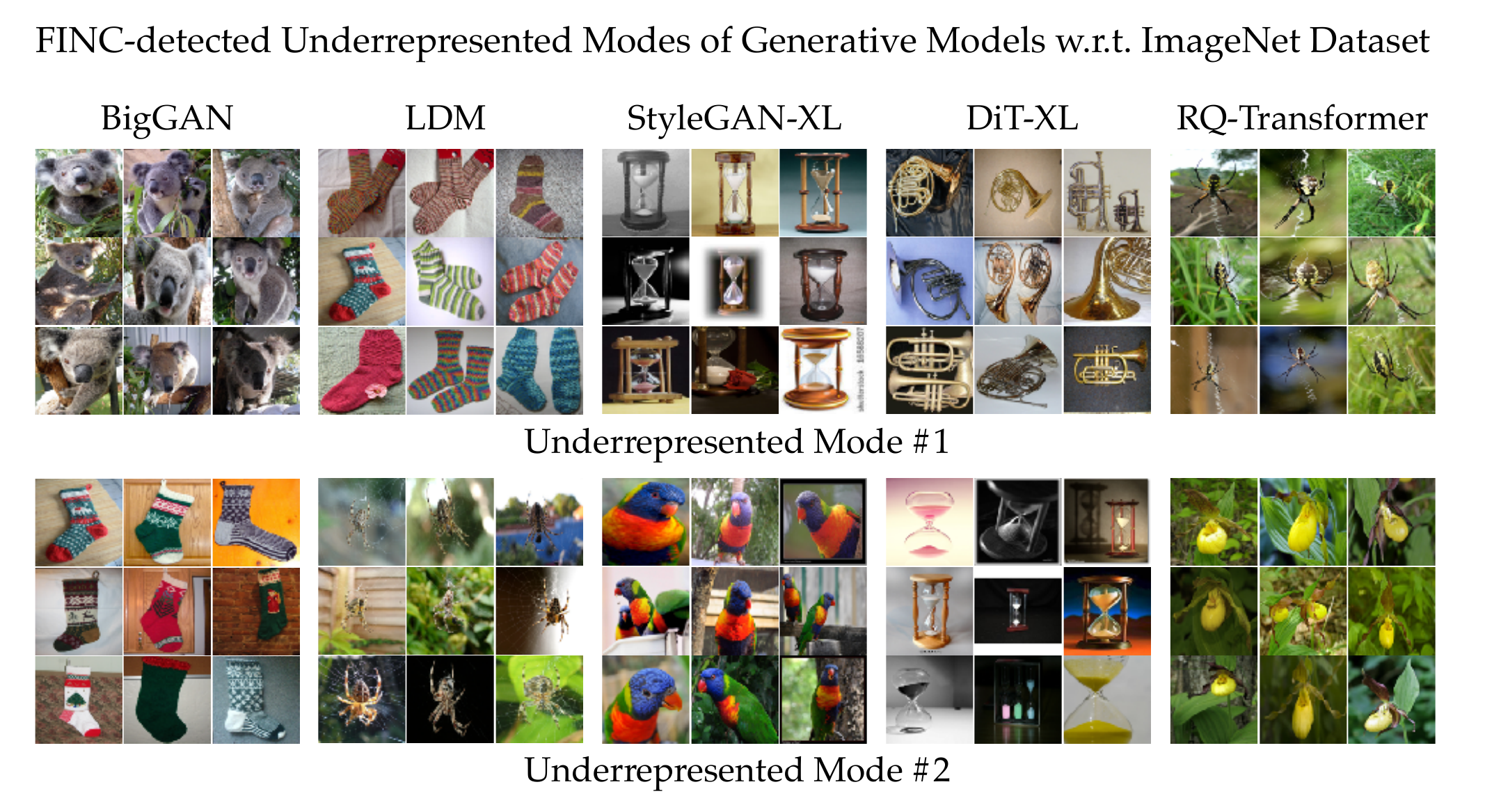}
    \caption{The FINC-identified top two underrepresented modes expressed by test generative models with a lower frequency than in the reference ImageNet dataset. Embedding is from DINOv2.}
    \label{fig:underrepresented_imgnet_dinov2}  
\end{figure}

\begin{figure}[t]
    \centering    
    \includegraphics[width=0.99\textwidth]{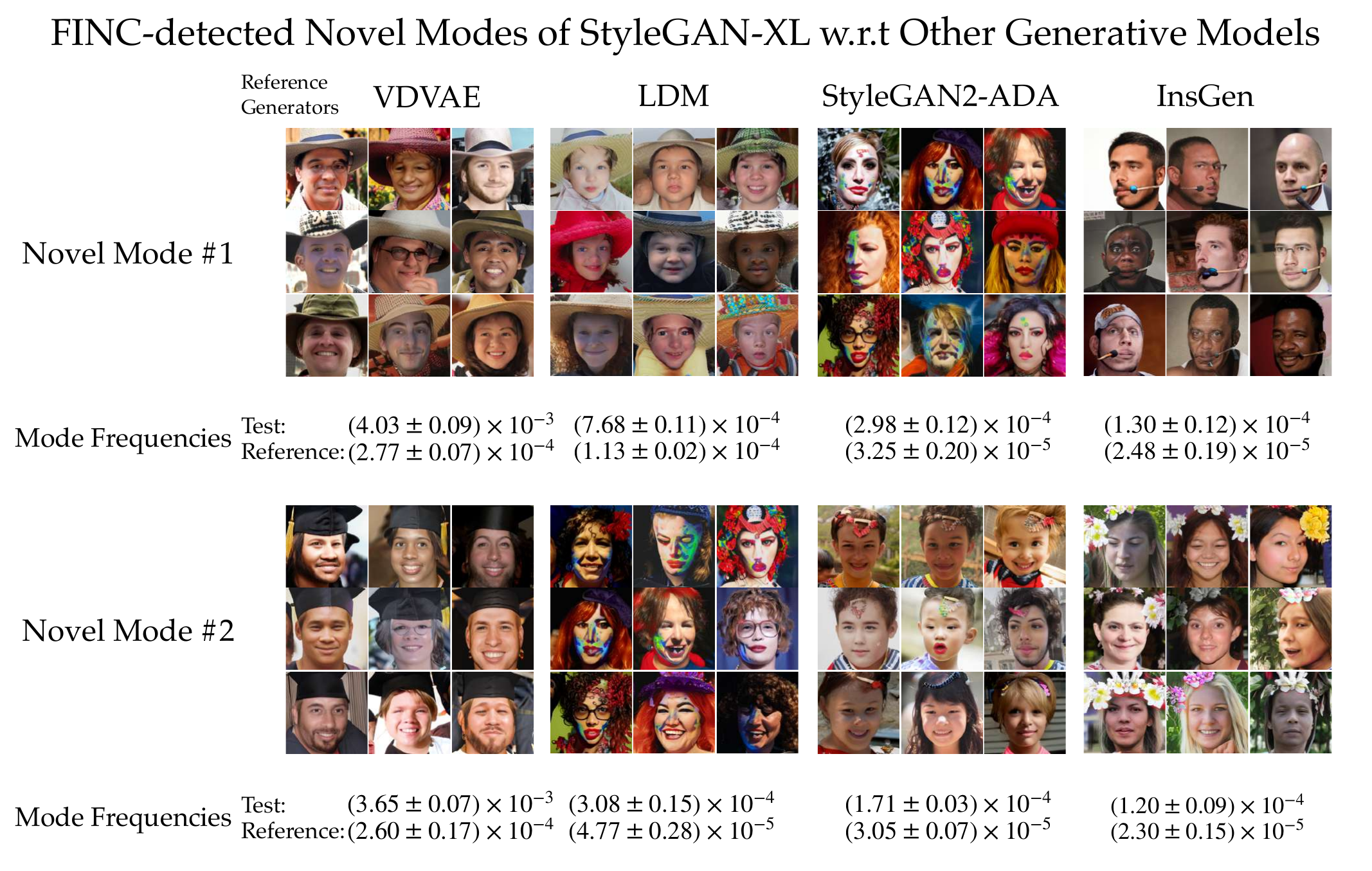}
    \caption{Representative samples from FINC-detected top-3 novel modes (novelty threshold $\rho=5$) of StyleGAN-XL compared to reference generative models.}
    \label{fig:novel_styleganxl_ffhq}  
\end{figure}

\end{document}